%% file: SRGPA_segmentation_tariel_08.tex
\documentclass[journal]{IEEEtran}
\usepackage[latin1]{inputenc}
\usepackage[dvips]{graphicx}

\usepackage{algorithm}
\usepackage{algorithmic}
\usepackage{color}
 \usepackage{amsmath,amsfonts,amssymb,stmaryrd}
\usepackage{multirow}

\begin{document}

\title{Conceptualization of seeded region growing by pixels aggregation. Part 4: Simple, generic and robust extraction of grains in granular materials obtained by X-ray tomography}

\author{Vincent Tariel}

\maketitle
\begin{abstract}
This paper proposes a simple, generic and robust method to extract the grains from experimental tridimensionnal  images of granular materials obtained by X-ray tomography. This extraction has two steps: segmentation and splitting. For the segmentation step, if there is a sufficient contrast between the different components, a classical threshold procedure followed by a succession of morphological filters can be applied. If not, and if the boundary needs to be localized precisely, a watershed transformation controlled by labels is applied. The basement of this transformation is to localize a label included in the component and another label in the component complementary. A "soft" threshold following by an opening is applied on the initial image to localize a label in a component.  For any segmentation procedure, the visualisation shows a problem: some groups of two grains, close one to each other, become connected.  So if a classical cluster procedure is applied on the segmented binary image, these numerical connected grains are considered as a single grain. To overcome this problem, we applied a procedure introduced by L. Vincent in 1993.\\
This grains extraction is tested for various complexes porous media and granular material, to predict various properties (diffusion, electrical conductivity, deformation field) in a good agreement with experiment data. 
\end{abstract}
\IEEEpeerreviewmaketitle

\section{Introduction}
The 3D reconstruction of microstructures by microtomography is based on
X-ray radiography. From several radiographs, obtained
after rotations of a specimen, and with an appropriate algorithm, 3D images
of the sample are obtained at a micro scale. This article opens the way to predict the physical properties using  3D images where each voxel belongs to a grain of the granular phase or to the matrix phase\footnote{This matrix phase can be void (granular A and D) or solid (granular B and C).}.
For example, 3D images of the microstructure can be introduced in a numerical
homogenization process by computer, in order to predict their macroscopic
physical properties from the microscopic properties. The first step to achieve this goal is to perform a correct grains extraction. Due to the acquisition of radiographs and to the numerical reconstruction, noise is apparent in the images, as well as some linear grey-level artefacts resulting in impressive images with generally a too weak quality for a quantitative and automatic use.\\
This paper proposes a simple, generic and robust method to achieve this goal. Simple means that this method can be used by anybody who is not a specialist of image processing. Generic means that this method can be applied in a wide range of materials. This method has been applied for granular materials (see figure~\ref{image}) but its extension to other materials is straightforward. Robust means that the extraction is few sensitive with a "little" variation of the parameters.\\
This method has two steps: 
\begin{enumerate}
\item segmentation: to transform the grey-level image into an image with different components,
\item splitting: to transform a granular component into a set of grains.
\end{enumerate}
1) Depending on the histogram shape, the segmentation is applied on the initial image to use the tint information or on the gradient image to use the boundary information as sharp variation of grey-level. In these both approaches, a succession of morphological filters is used:
\begin{itemize}
\item after the threshold, to remove the small islands and to fill the holes of islands for the tint information,
\item before the watershed transformation, to localize two labels: one included in the component, the other included in the component complementary for the boundary information.  
\end{itemize}
2) Whatever the approach, the final result shows one difficulty:  some groups of two
grains close one to each other become connected after the segmentation. So if a classical cluster procedure is applied on the segmented binary image, these numerical connected grains are considered as a single grain.  To overcome this problem, a splitting procedure is applied to separate the connected grains.\\
The outline of the rest of the paper is as follows: in Sec.~II, the images characteristics obtained by X-ray tomography and the different materials are explained. In Sec.~III, the classical threshold procedure is presented. In Sec.~IV, the watershed transformation controlled by labels and the localization labels are explained. In Sec.~V, a splitting procedure is applied to individualize each grain in the granular component. In Sec.~VI, concluding remarks are made. 

\input{material}

\input{threshold}
\input{label_watershed}
\input{dynamic}



\section{Conclusion}
X-ray tomography images involve specific problems of image segmentation, due
to some artefacts generated during the reconstruction. In the case of
granular materials, we have seen that many problems can appear during the
segmentation. In the first part of this paper, we presented the classical  threshold segmentation. To overcome the problem due to the low brightness between the components, we introduced a simple, generic, robust method of segmentation, called double labels watershed. This method is simple because the chain of operator is always the same, generic because this method has been applied in a wide range of material, robust because some "little" fluctuation of the parameters does not change the final segmentation.  Then to split the connected grains generated by the segmentation in tridimensionnel image, we have applied a classical procedure introduced  by L. Vincent. \\
 This simple, robust and generic segmentation can be applied efficiently in using the framework introduced in the previous articles. The initial goal, to give a method for any user not especially specialist in image processing, is reach. 

\section*{Acknowledgment}
I would like to thank my Ph.d supervisor, P. Levitz, for his support and his trust. The author is indebted to E. Gallucci, D. Jeulin for valuable discussion and C. Wiejak for critical reading of the manuscript.  I express my gratitude to the Association Technique de l'Industrie des Liants Hydrauliques (ATILH) for its financial support and the French ANR project "mipomodim" No. ANR-05-BLAN-0017 for their financial support.

\bibliographystyle{plain}
\bibliography {../bibliogenerale}
\end{document}

%% file: material.tex
\section{Materials and methods}

\subsection{Microtomography}
\begin{figure}
\begin{center}
\includegraphics[width=4cm]{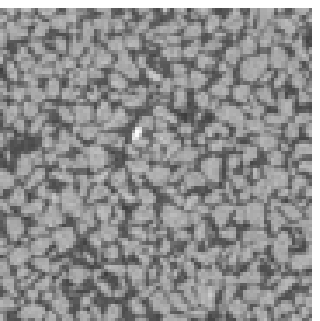}\includegraphics[width=4cm]{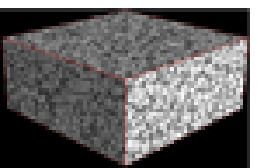}
\includegraphics[width=4cm]{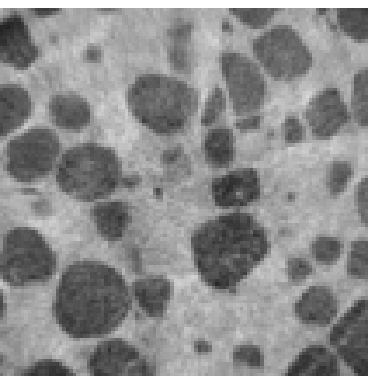}\includegraphics[width=4cm]{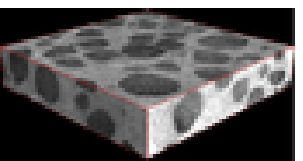}
\includegraphics[width=4cm]{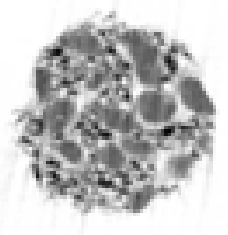}\includegraphics[width=4cm]{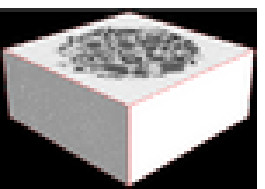}
\includegraphics[width=4cm]{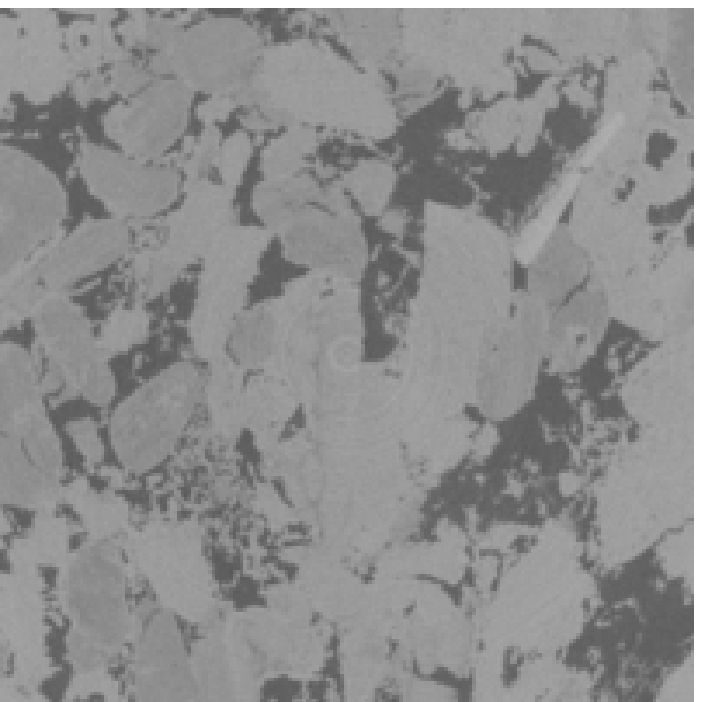}\includegraphics[width=4cm]{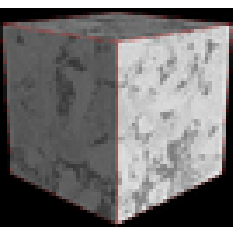}
\end{center}
\caption{
Figure 1: Granular Material A, size=450x450x200,  resolution=14 microns; figure 2: Granular Material B, size=500*500*100, resolution=3 microns; figure 3: Granular Material C, size:350x350x150, resolution=3 microns,  there are three components: the matrix in light and two grain classes: small  dark grains and large grains with a medium average grey level ;  figure 4: granular material D, size=700x700x700 resolution=3 microns, there are three components: the void in dark and two grain classes: one lighter and the other with a medium average grey level.}
\label{image}
\end{figure}

Microtomography is a non-destructive 3D-characterisation technic providing a three-dimensional image\footnote{Each image is a \textit{rectangular}  of the space  $E$. $E$  is a 3-dimensional discrete space $\mathbb{Z}^n$, consisting of lattice points which coordinates are all integers. The elements of a 3-dimensional image array are called voxels.}. Each voxel of the image is associated to a cube included in the material, under investigation\cite{Bernard2005}. In first order, its grey-value is the space average of linear X-ray absorption coefficient of the different solids and fluids contained into it. But since more often the tomographic reconstruction amplifies the noise of the projections, and generates artefacts, there is extra-term given impressive images with generally a too weak quality for a quantitative and automatic use. Also, the materials are different in the chemical composition and in the geometrical organisation (see figure~\ref{image}).
Due to the materials variety and the images defects, a generic, simple and robust segmentation procedure has been developed.
\subsection{Materials and applications} 
For the granular A, the segmented data come from a mechanical triaxial test on a sand specimen realised under a synchrotron   microtomograph (ESRF, ID15A) to follow the structural evolution of the granular media. Digital Image Correlation is used to observe and detect the strain localisation mechanisms at the grain scale\cite{Hall2008,Lenoir2008}. This work is funded by the French project ANR-05-BLAN-0192 (J. Desrues project coordinator).\\
For the granular B and C, the segmented data come from pyrotechnical specimen realised  by Laboratoire 3S-R, Grenoble  under a laboratory microtomograph designed by  Skyscan$^\circledR$. The finite element code is used  to compute stresses and strains as well as other fields like the thermal flux and the temperature distribution in the material at a micro scale in using \cite{Jeulin2007}. This work is funded by les Mines de Paris and le Centre d'Etudes de Gramat (DGA). \\
For the granular D, the segmented data come from estaillade limestone realised under a laboratory microtomograph designed by  Skyscan$^\circledR$ to understand  the effects of the porous structure in resistivity index curves\cite{Han2008}.
\subsection{The image size} 
The segmentation procedure must be efficient because the size of the images obtained by X-ray tomography is huge. The previous articles \cite{Tariel2008,Tariel2008c,Tariel2008d} enable to improve the efficiency of the algorithms used for this prupose. Typically, for the biggest image of this article, the granular D with a size equal to 700*700*700=0.35 Giga voxels, the segmentation requires less than 6 hours of executing time with an Intel(R) Xeon(R) CPU 3.00GH and the allocation of RAM is 16 GO. This reasonable time makes the practical use  manageable on conventional computers. The algorithms are developed and implemented on the Open source software, called Population available at http://pmc.polytechnique.fr/$\sim$vta/Pop.tar
\subsection{Images characteristics} 
For every image, the grey-level is coded on one byte (0-255) and a median filter has been applied to minimize the ring artefact and to smooth in keeping the sharpness of the boundary. The images have an uniform illumination and   each component on the image has a specific brightness\footnote{This last assumption is not always verified. For example, the large grains with a medium average grey level in the granular B is divided into two components which  chemical composition is different and which linear X-ray absorption coefficient is the same. Without more information, we consider these two components as one component.}.  For the visualization convenience, the results are sometimes presented in 2D but the method has been applied efficiently in 3D for all materials.

%% file: threshold.tex
\section{Segmentation}

For each material, depending on the histogram shape, the classical threshold segmentation can be applied to extract a component, using tint information (1). If the contrast between the component and the background is low and if the boundary has to be well localized, the watershed transformation controlled by labels is applied  using the boundary information (2).\\
For the both approaches, a combination of morphological filters has to be applied in order to:
\begin{enumerate}
\item match the visual segmentation for (1) (the combination is an opening followed by a closing),
\item localize two labels for (2)  (the combination  is just an opening).
\end{enumerate}
This section is decomposed into two parts: threshold segmentation using tint information and watershed transformation using boundary information.

\subsection{Threshold segmentation using tint information}

\subsubsection{Threshold operation}
\begin{figure}
\begin{center}
\includegraphics[width=3.5cm]{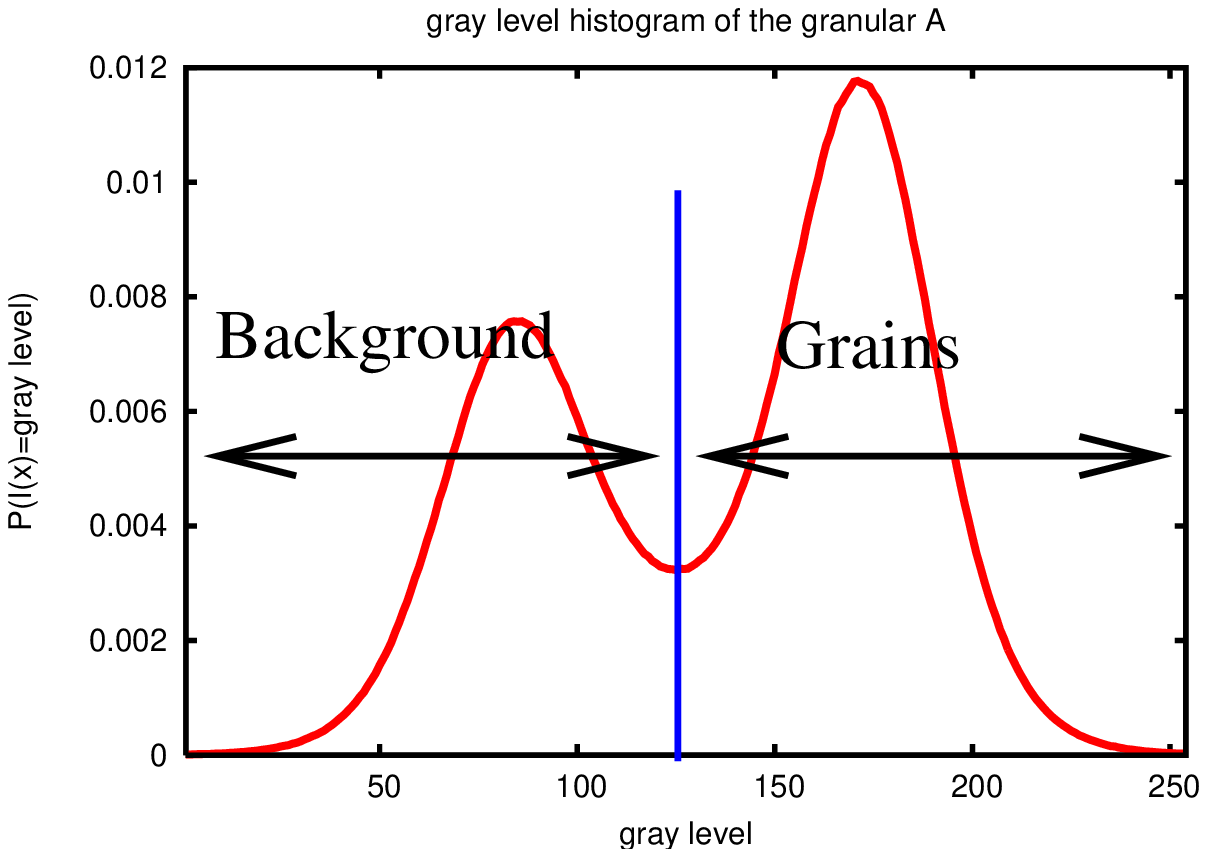}
\includegraphics[width=2.5cm]{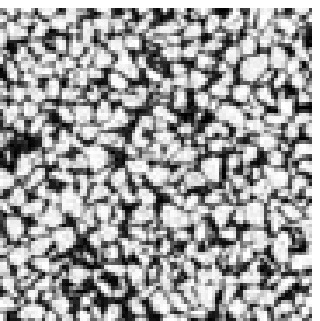}
\includegraphics[width=2.5cm]{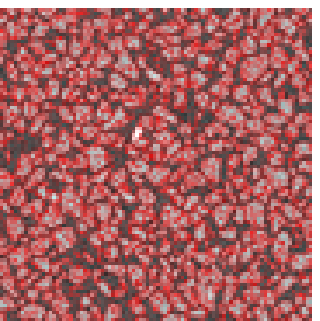}
\includegraphics[width=3.5cm]{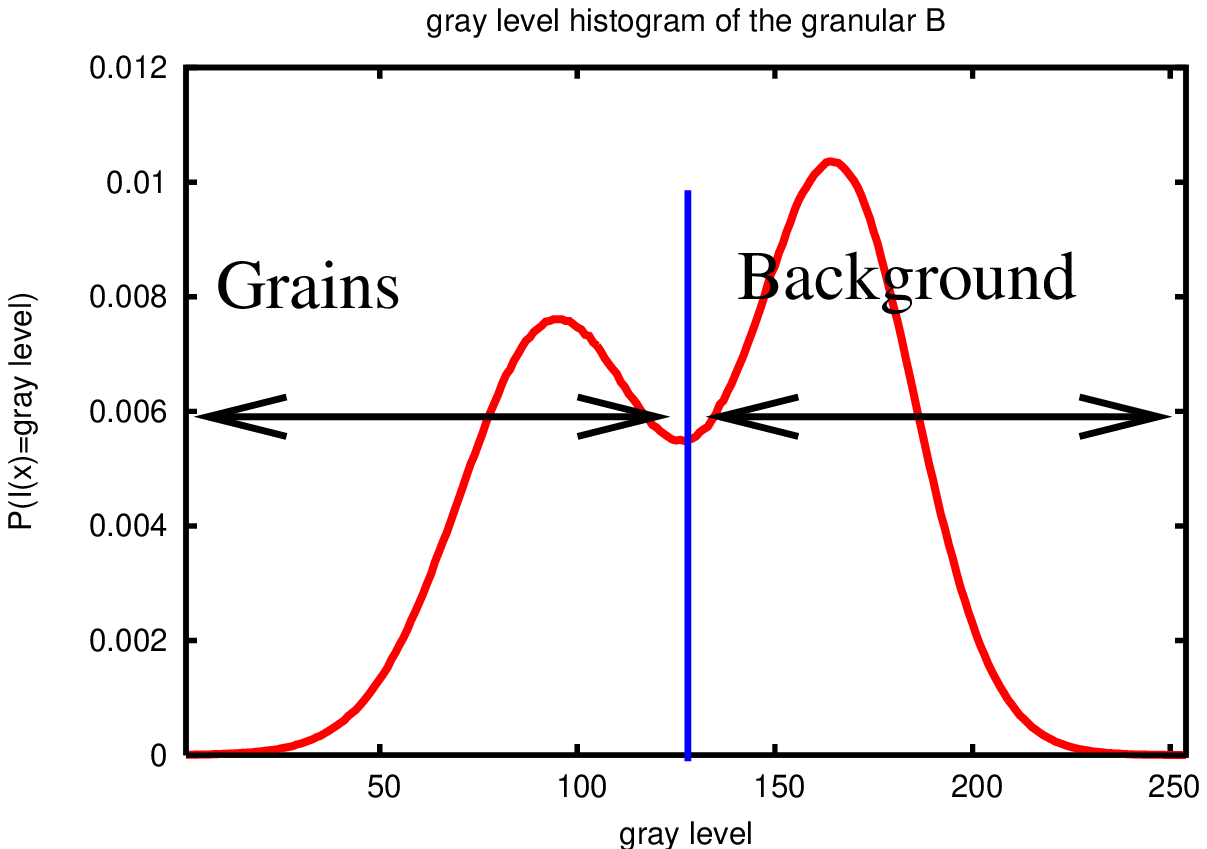}
\includegraphics[width=2.5cm]{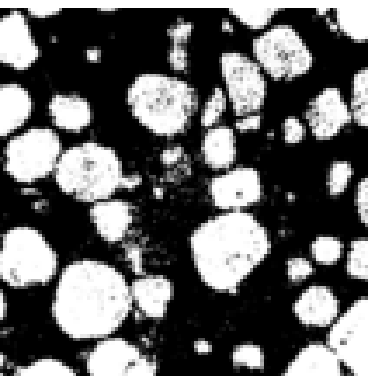}
\includegraphics[width=2.5cm]{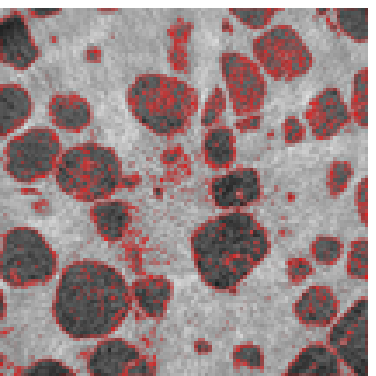}
\includegraphics[width=3.5cm]{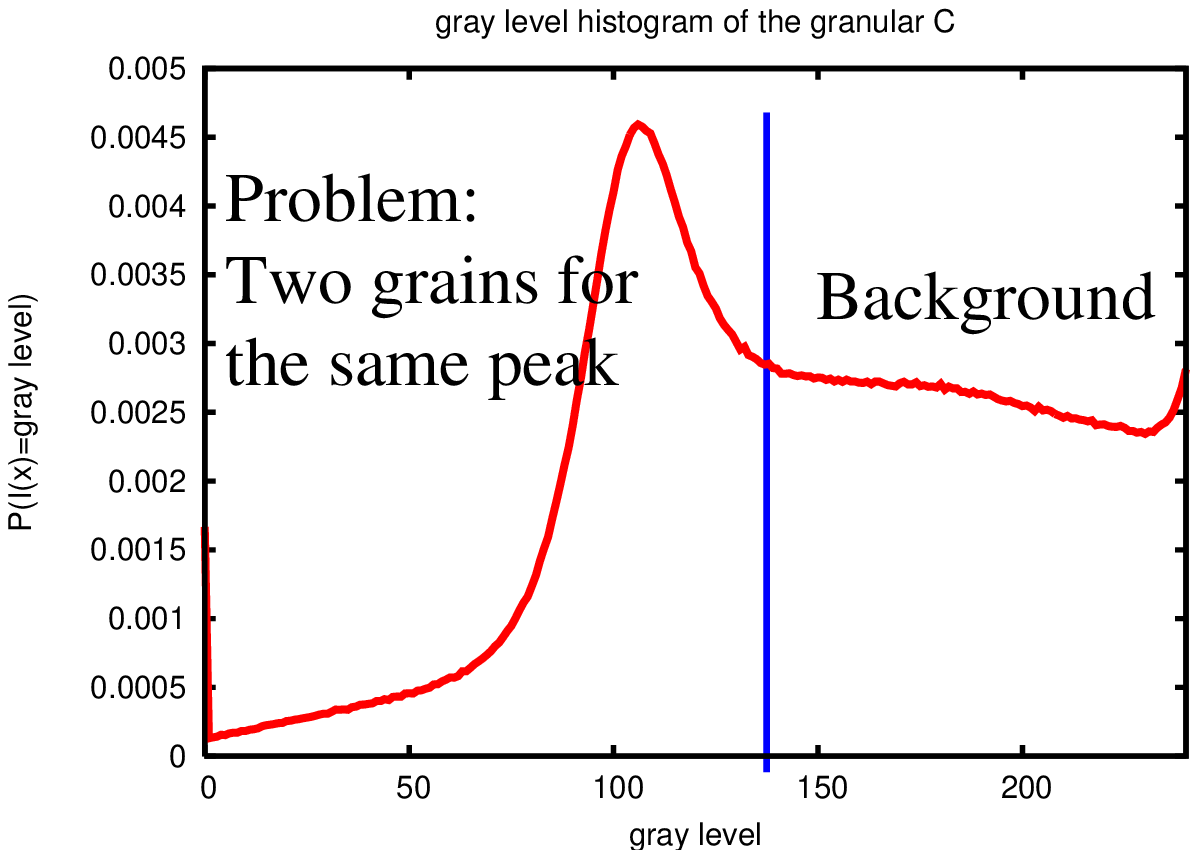}
\includegraphics[width=2.5cm]{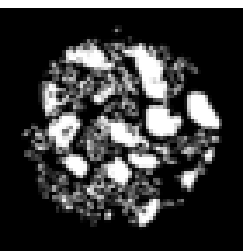}
\includegraphics[width=2.5cm]{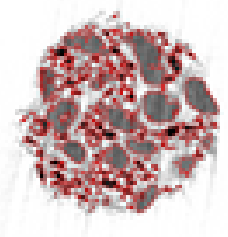}
\includegraphics[width=3.5cm]{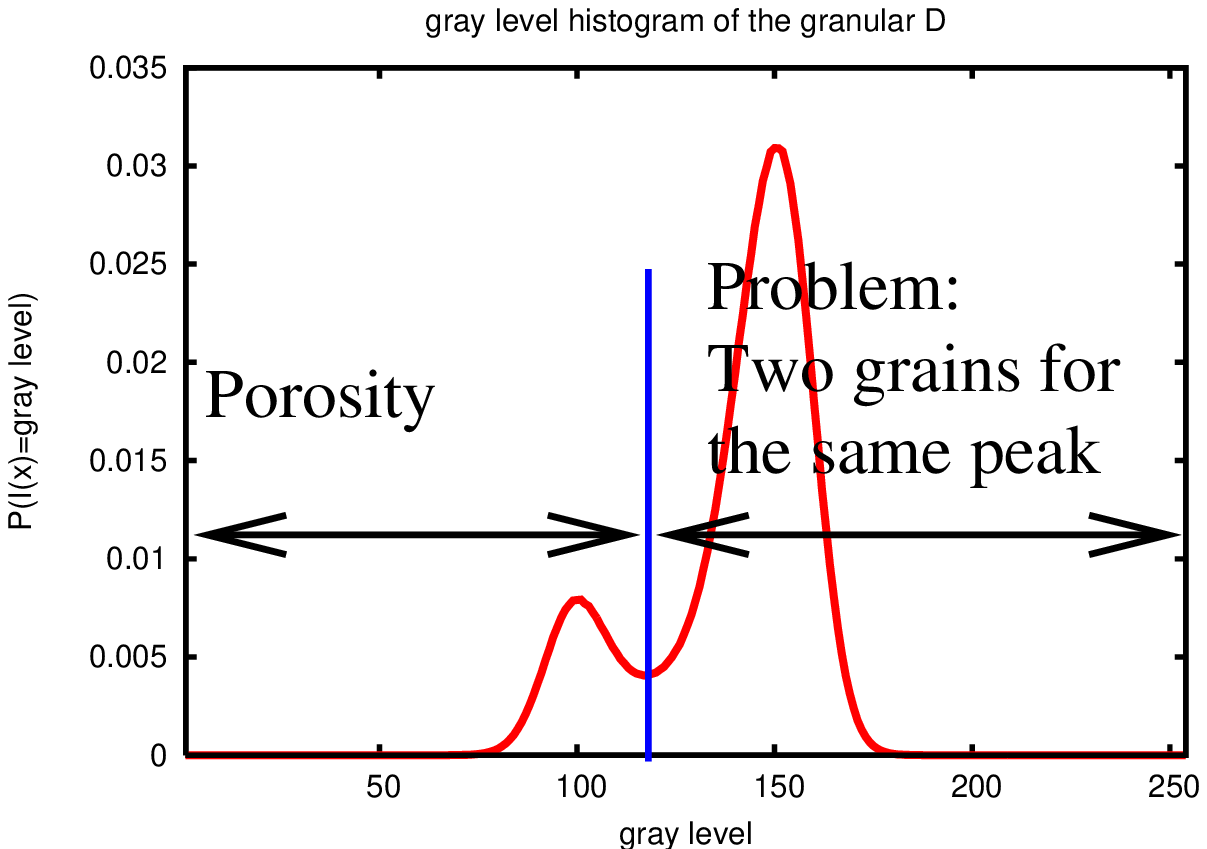}
\includegraphics[width=2.5cm]{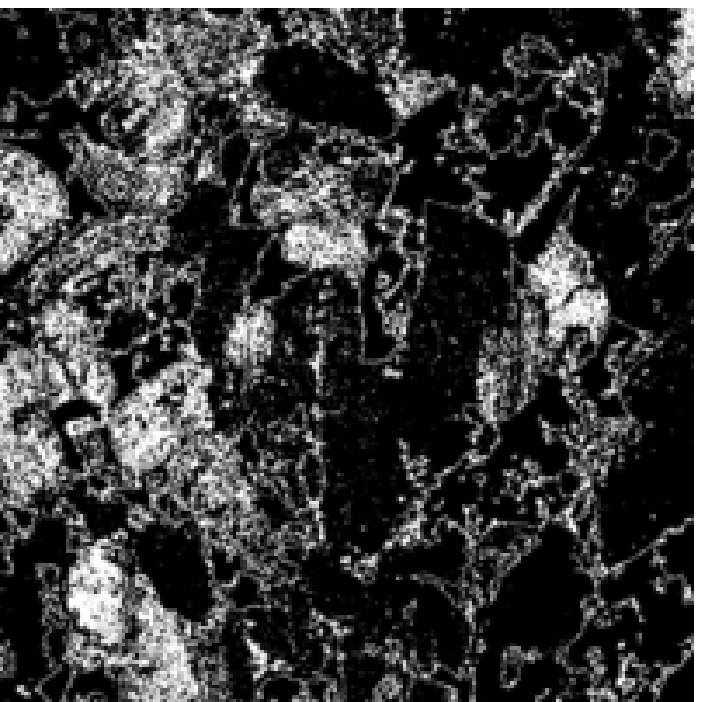}
\includegraphics[width=2.5cm]{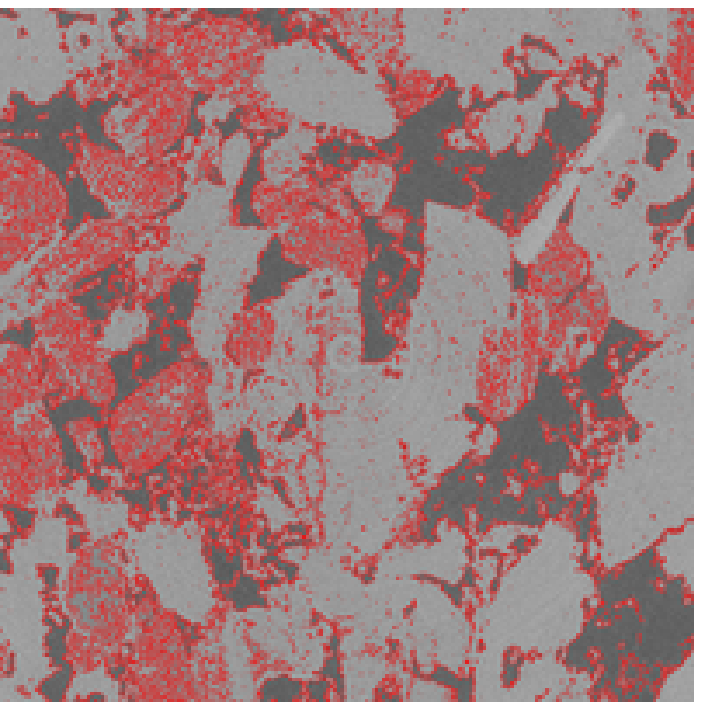}
\caption{Each serie is associated to a granular, the first image is the histogram, the second figure is the binary image after threshold, the third figure is the foreground of the binary image inner boundary on the initial image. \underline{Serie 1)}: granular A,  the range of the thresold is 125:255, despite some holes in the numerical grains and some voxels outside the grains, the numerical segmentation is quite similar to the visual segmentation,  \underline{serie 2)}: granular B, the range of the thresold is 0:125. \underline{Serie 3)}: granular C, the right mode is populated with voxels that form the two grains classes  within the image. We have selected arbitrarily the range 120:145 to extract the darkest grains.  The binary image shows some clusters of voxels corresponding to the middle grey grains, some lines corresponding to the boundary of the porosity (the darkest phase) due to a halo artefact and some isolated voxels corresponding to the light grains. \underline{Serie 4)}: granular D, the range is 80:144, idem that granular C}
\label{histo}
\end{center} 
\end{figure}

Given that each component has a specific brightness, the threshold operation uses this contrast information to extract the components.  Threshold selection is usually based on the information contained in the grey level histogram of the image (see figure~\ref{histo}).\\ 
The correspondence between the component and a mode in the histogram is required for threshold segmentation. 
A certain threshold range is chosen. The label '1' is assigned to each image voxel, which gray-level belongs to this range, and '0' label is assigned to each image voxel which gray-level does not belong to this range. The threshold range is selected to best separate the mode of the histogram, see \cite{Rosenfeld1982,DAVIS1975,PANDA1978,WESZKA1978}. Manually, the range is chosen on the valleys of the histogram (see figure~\ref{histo}). 
For the granular A and B, the binary image seems like the real grains aside some holes and isolated voxels. The next paragraph explains how to correct these defects. For the granular C and D, the binary image is very different of the real grains. The reason is that one mode in the histogram is populated with voxels that form the two classes of grains because the contrast between these two classes is low. For these two granular materials, it is impossible to filter the binary image to get a good segmentation.  The subsection~\ref{sec:watershed} gives a method to treat this task.

\subsubsection{Morphological filtration}
Briefly, the four basic operator of the mathematical morphology are presented\cite{Serra1982}.\\
Erosion of object $A$ by the structural element $B$ is defined by\footnote{the inner boundary using for visualisation is defined as: $\partial A = A \setminus( A\ominus B) $}:
\[
A \ominus B = \{z | (B)_{z} \subset A\}
\]
Dilatation of object $A$ by the structural  element $B$ is defined by:
\[
  A \oplus B = \{z | (B)_{z} \cap A\neq\emptyset\}
\]
The Opening of $A$ by $B$ is obtained by the erosion of $A$ by $B$, followed by dilatation of the resulting structure by $B$:
\[
A \circ B = (A \ominus B) \oplus B 
\]
The Closing of $A$ by $B$ is obtained by the dilatation of $A$ by $B$, followed by erosion of the resulting structure by $B$:
\[
A \bullet B = (A \oplus B) \ominus B 
\]

\begin{figure}
\begin{center}
\includegraphics[width=2.1cm]{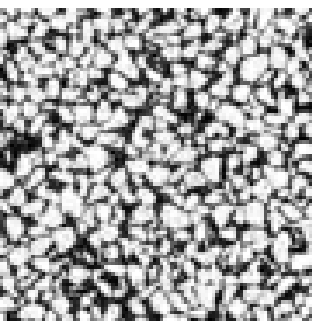}
\includegraphics[width=2.1cm]{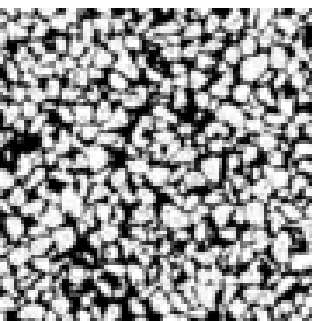}
\includegraphics[width=2.1cm]{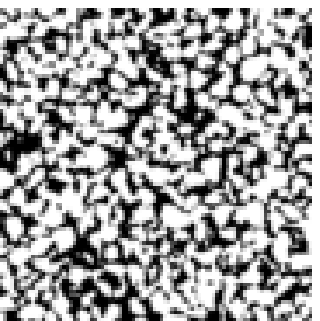}
\includegraphics[width=2.1cm]{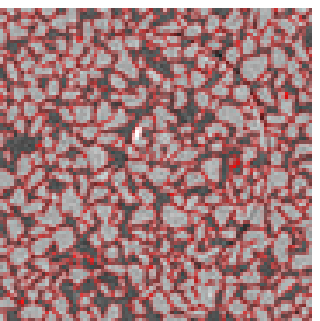}\\
\includegraphics[width=2.1cm]{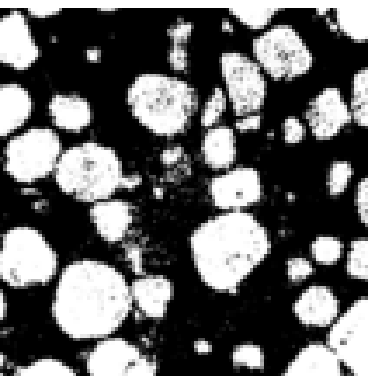}
\includegraphics[width=2.1cm]{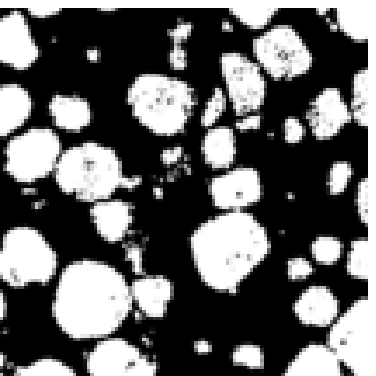}
\includegraphics[width=2.1cm]{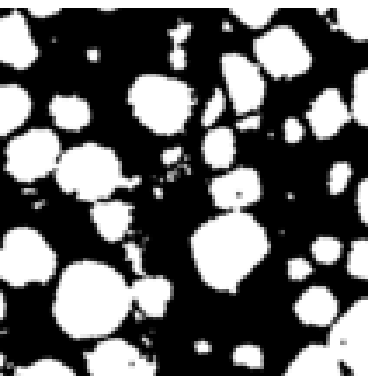}
\includegraphics[width=2.1cm]{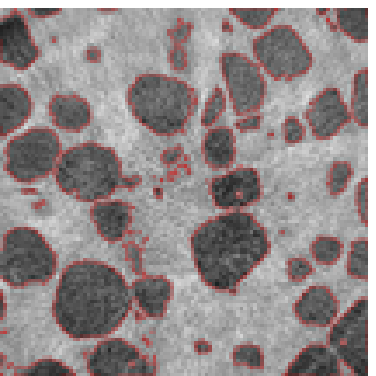}
\caption{For both serie, the binary image is the application of threshold, the second image is the application of opening on the first image to remove the isolated voxels, the third image is the application of closing on the second image to fill the holes, last image: foreground of the inner boundary of the third image in red on the initial image.}
\label{filter}
\end{center} 
\end{figure}

As the grains are isometric for all granular materials, the structural element is chosen isometric. The structural element is associated to the 26-connectivity in the cubic grid.\\
To remove the isolated voxels for keeping the clusters size, the opening is applied. To fill the holes for keeping the clusters size, the closing is applied. Let $A$, the binary image after the threshold application. The filtration is only: $ (A \circ B) \bullet B$ (see figure~\ref{filter}). We have a good agreement between the visual segmentation and the numerical segmentation but first the numerical boundary does not match closely the visual segmentation, second some groups of two grains close one to each other become connected after the segmentation. So, the following method has to be applied if
\begin{enumerate}
\item for the purpose of the application, it is necessary to have a good match between the numerical boundary and the "real" boundary,
\item the contrast between the components is low.
\end{enumerate}

%% file: label_watershed.tex
\subsection{Watershed transformation using boundary information}
\label{sec:watershed}
\subsubsection{Double labels watershed}
\begin{figure}
\begin{center}
\includegraphics[width=4cm]{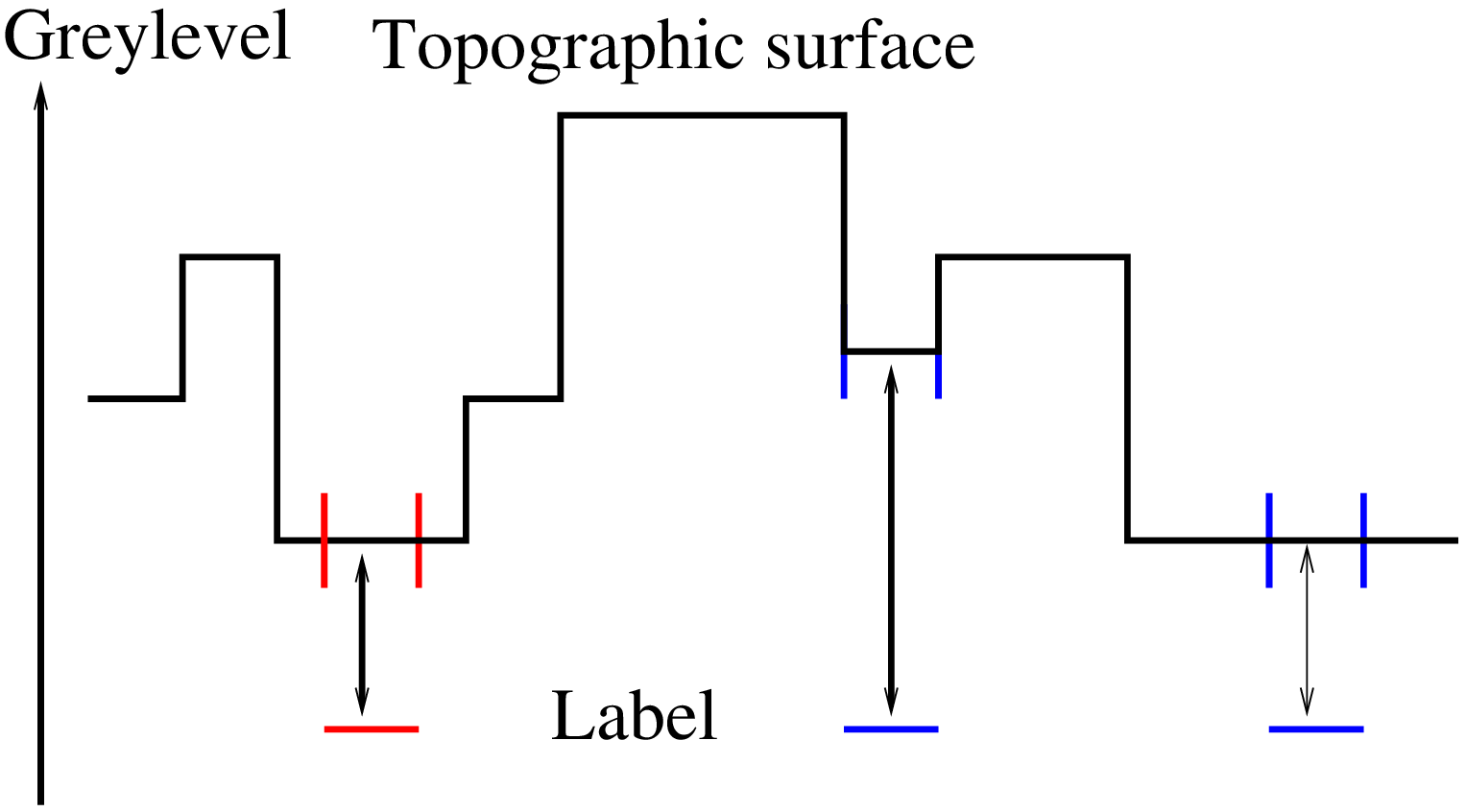}
\includegraphics[width=4cm]{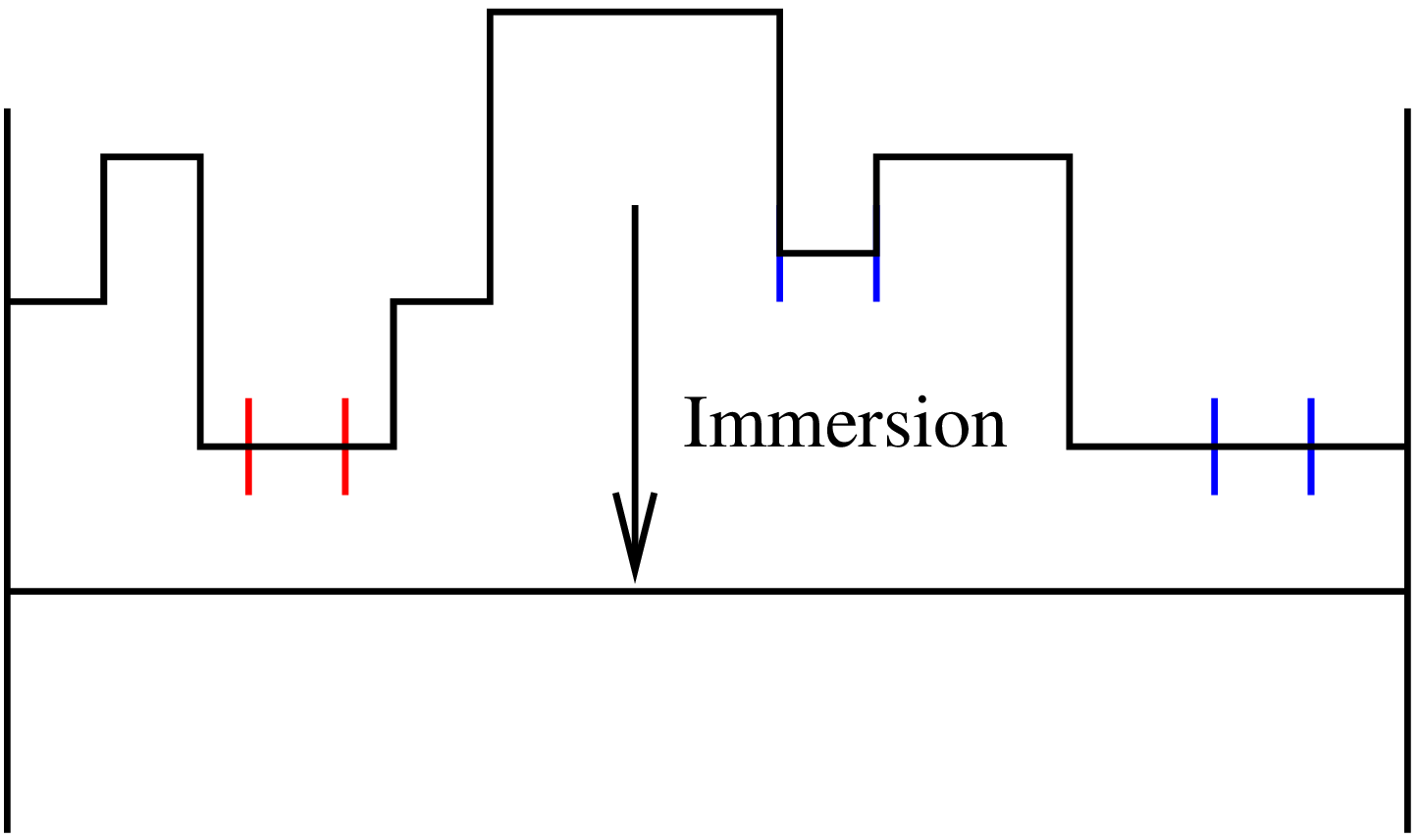}
\includegraphics[width=4cm]{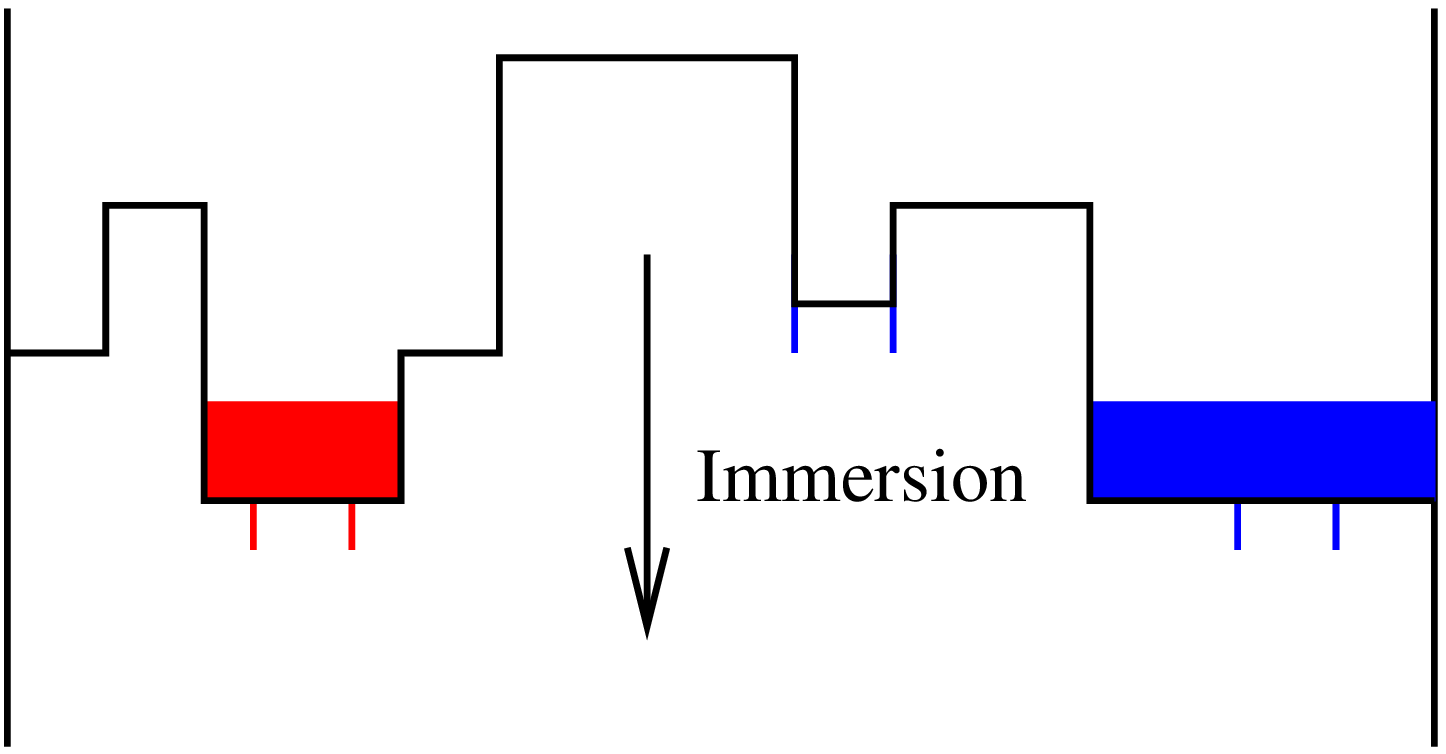}
\includegraphics[width=4cm]{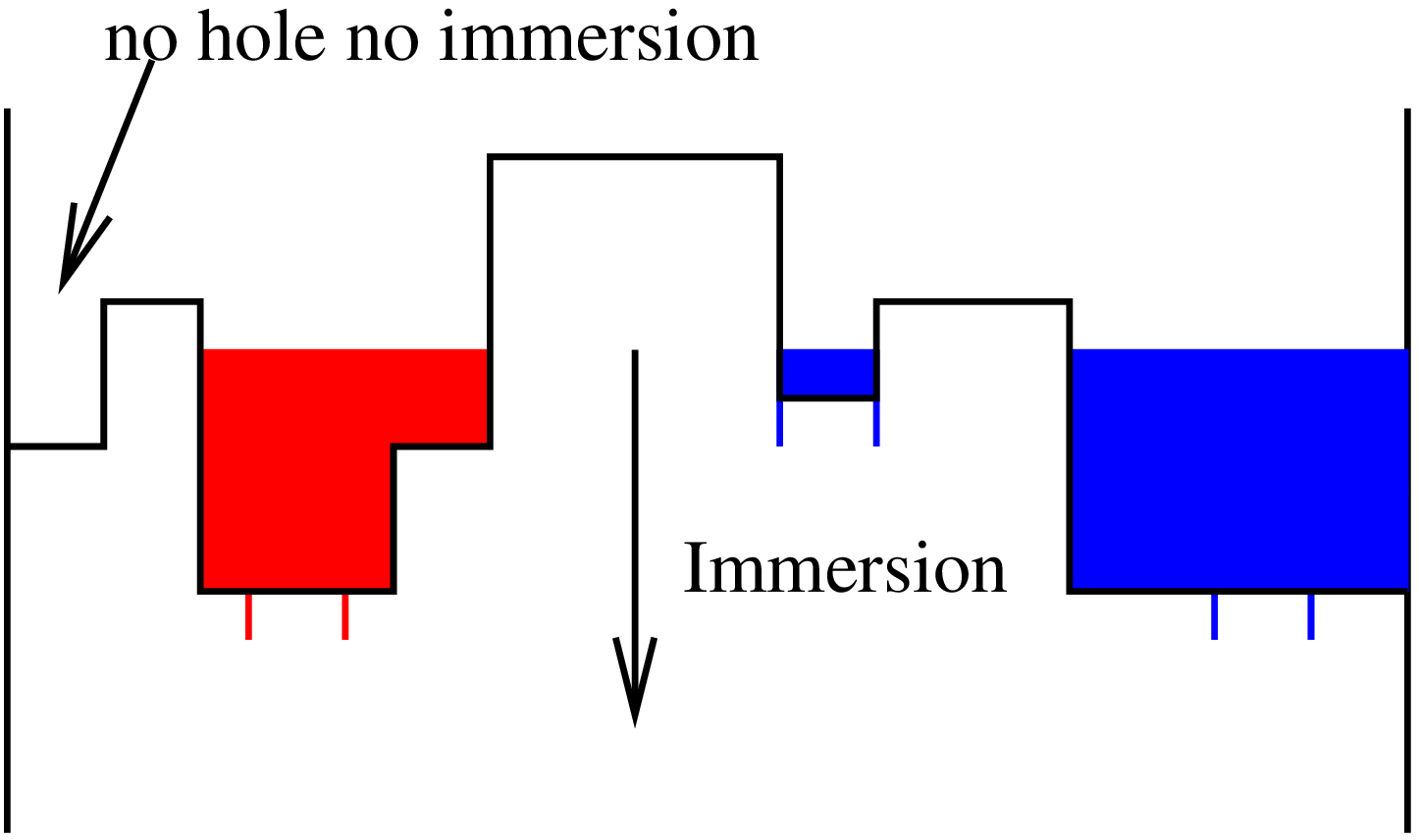}
\includegraphics[width=4cm]{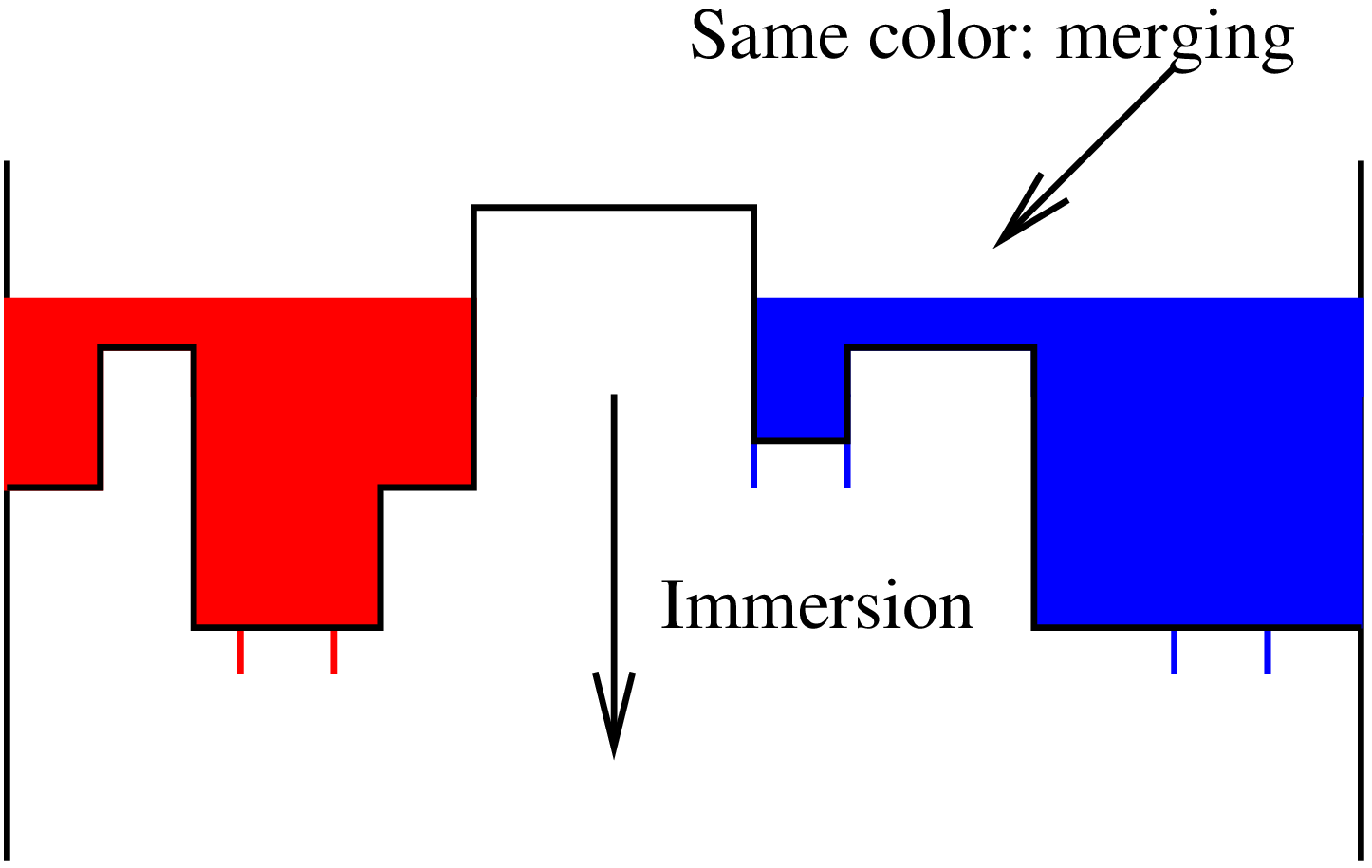}
\includegraphics[width=4cm]{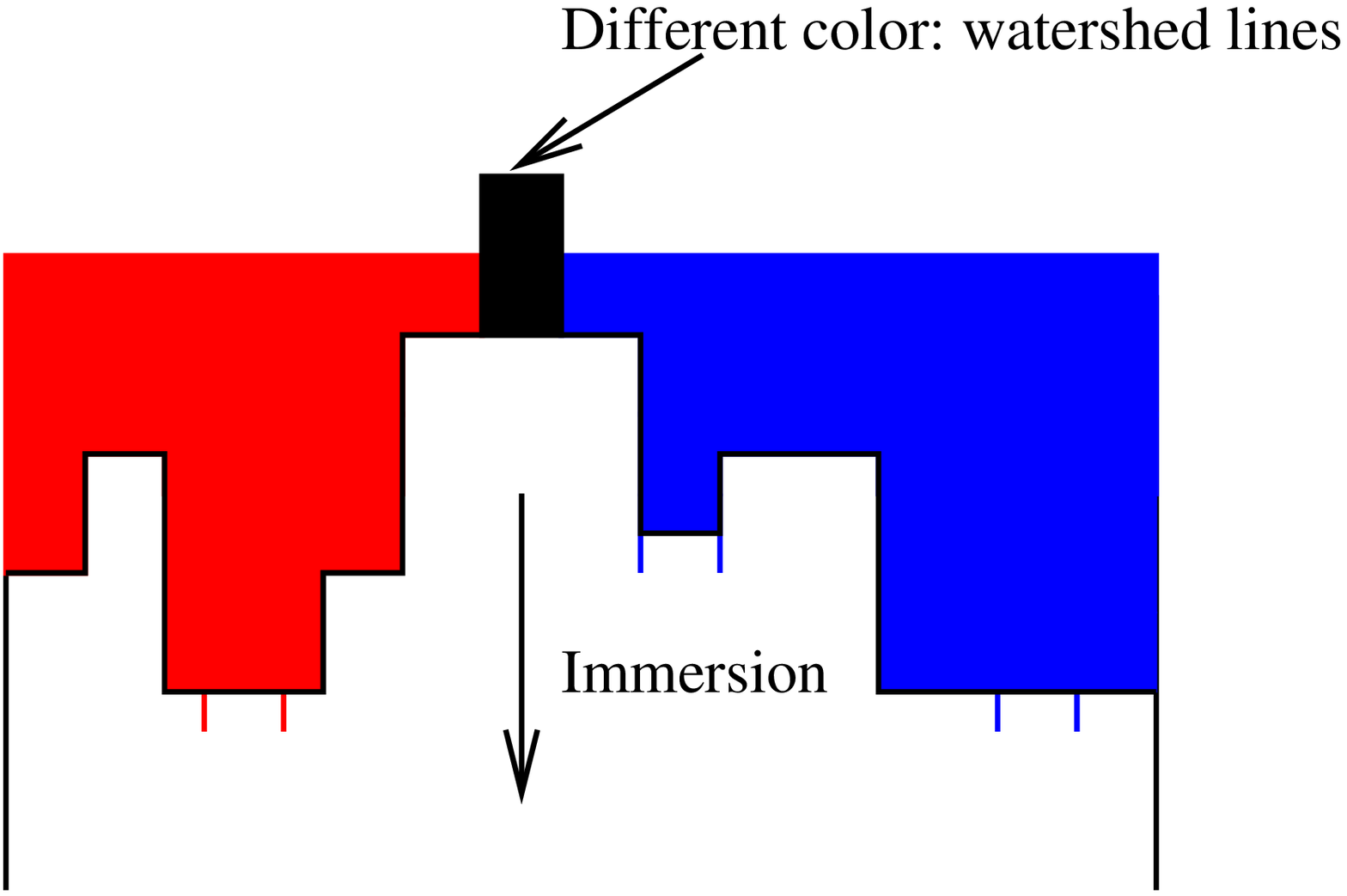}
\caption{This transformation requires two images: the topographic surface (a grey-level image) and the label image. The process is: 1) association of each label to a hole 2) immersion 3) the water enters in the topographic by the holes and the catchment basins take the colour of the hole, 4) a part of the topographic is not merged although its level is under the level of the immersion, 5) fusion of two catchment basins of same colour, 6) creation of dam when two catchment basins have different colours. \textit{A video is available at http://pmc.polytechnique.fr/$\sim$vta/water.mpeg}  }
\label{water}
\end{center} 
\end{figure}

\begin{figure}
\begin{center}
 \includegraphics[width=10cm]{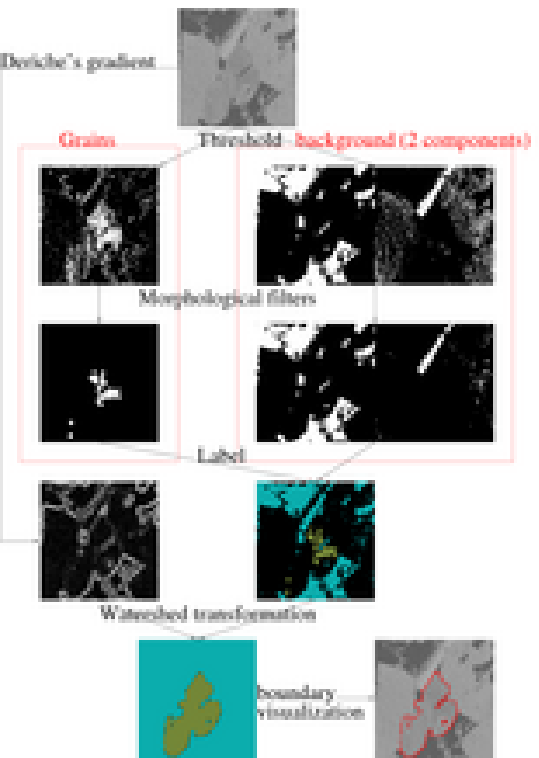}
\caption{Starting from the initial image, two labels are localized: one included in the component and the other in the component complementary. To localize a label in a component, a threshold followed by opening is operated. Since the complementary is composed by n components (2 in this example), the localization is done for each component such as the final localization for the complementary is the addition of these specific localizations. When a watershed transformation is applied on the gradient with these two labels, the catchment basins are well localized on the grains. There is a good match between the visual segmentation and the numerical segmentation.}
\label{waterlabel}
\end{center} 
\end{figure}

An efficient segmentation procedure developed in mathematical morphology is the watershed segmentation  \cite{Beucher1979}, usually implemented by a flooding process from labels.
 We recall these tools in this section, based on our application for the segmentation of the 3D images of materials.\\
\emph{Minima watershed:} Any greyscale image can be considered as a topographic surface and all boundaries as sharp variations of the grey level. When a gradient is applied to an image,  boundaries are enhanced. When the topographic surface obtained from the gradient is flooded from its minima, the waterfronts meet on watershed lines in 2D, and on watershed surfaces in 3D. A partition of the investigated volume is obtained, where the catchments basins are separated by the watershed surfaces.  However, in practice, this merging produces an important over-segmentation due to noise or local irregularities in the gradient image, generating a set of uncontrolled and unwanted markers. To avoid this problem coming from too many minima, the image, $I$, is usually  filtered. It is a composition of vertical (with respect to the grey level) and horizontal filters, in order to individualize each grain with a single marker. This individualisation step is complicated\cite{Tariel2008}. It is the reason why the labels controlled watershed is used\cite{Beucher1991}.\\
\emph{Labels controlled watershed}: The labels controlled watershed is similar to the minima watershed, beside a catchment basin is associated to a label (see figure~\ref{water}).
\\
\emph{Component by component :} Since the materials are different in the chemical composition and in the geometrical organisation, it is difficult to extract each component in the same procedure. It is simpler to extract component by component. Starting from the simplest component to extract, we proceed to extract the next simplest component each time. Notice that the last component is easily extracted because it is the complementary of the other components addition.\\
 Except the last component, the extraction procedure is:
\begin{enumerate}
\item  to localize two labels: one included in the component and the other in the  component complementary (the next paragraph is dedicated to this task),
\item  to apply the Deriche's operator\cite{Deriche1987} on the initial image to get the gradient image,
\item  to apply the  watershed transformation controled by labels on this gradient image with these labels (see figure~\ref{waterlabel}).
\end{enumerate}
The component is the catchment basin associated to the label included in it   
\subsubsection{Localize a Label in a component}
\begin{figure}
\begin{center}
\includegraphics[width=2cm]{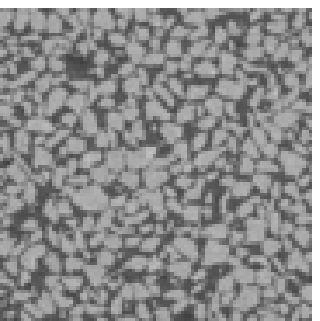}
\includegraphics[width=2cm]{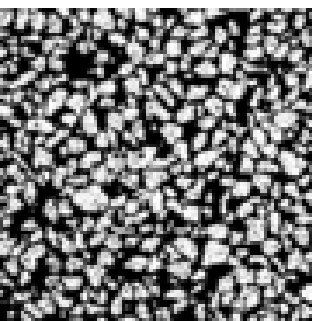}
\includegraphics[width=2cm]{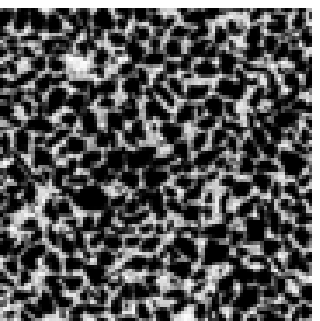}\\
\includegraphics[width=2cm]{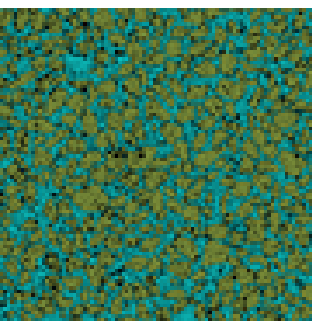}
\includegraphics[width=2cm]{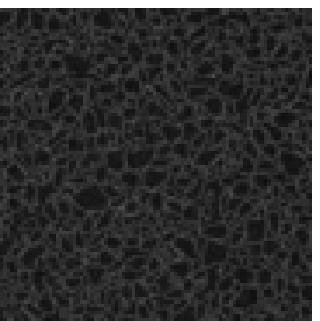}
\includegraphics[width=2cm]{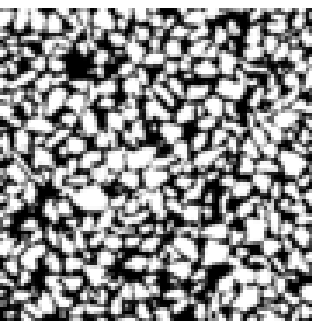}
\includegraphics[width=2cm]{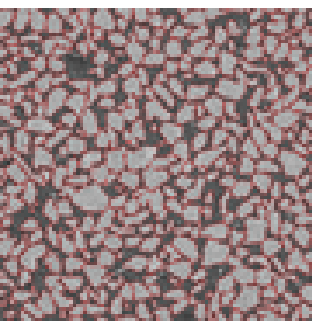}\\
\includegraphics[width=2cm]{image/iex.eps}
\includegraphics[width=2cm]{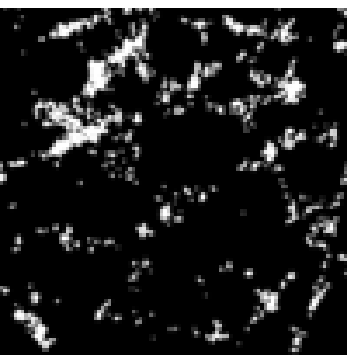}
\includegraphics[width=2cm]{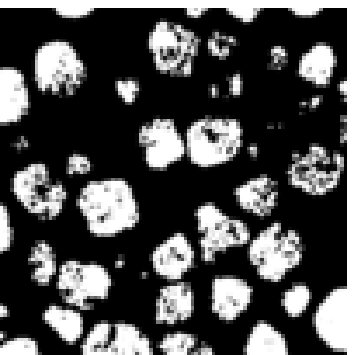}\\
\includegraphics[width=2cm]{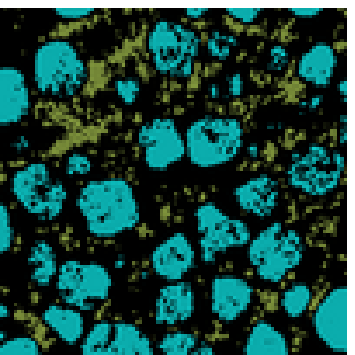}
\includegraphics[width=2cm]{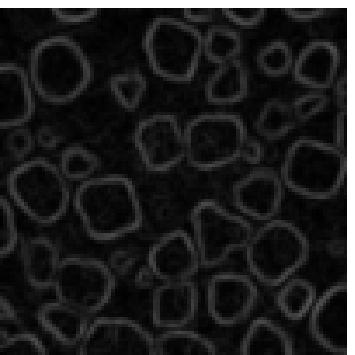}
\includegraphics[width=2cm]{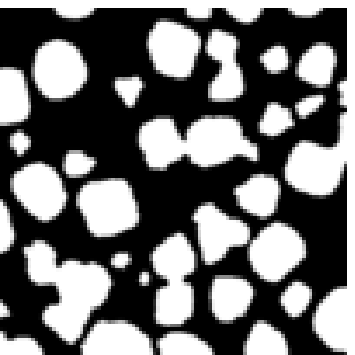}
\includegraphics[width=2cm]{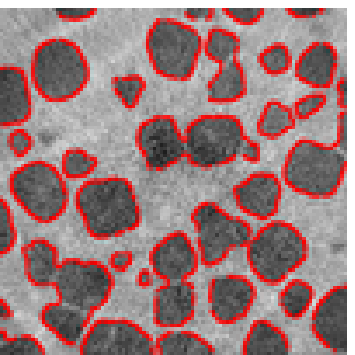}\\
\includegraphics[width=2cm]{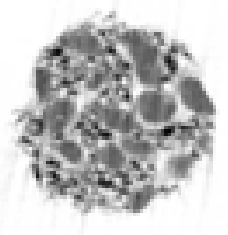}
\includegraphics[width=2cm]{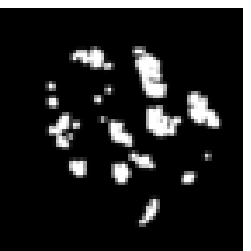}
\includegraphics[width=2cm]{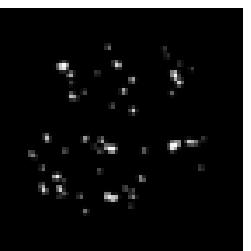}
\includegraphics[width=2cm]{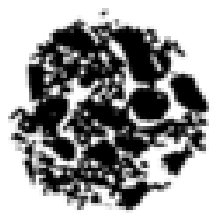}\\
\includegraphics[width=2cm]{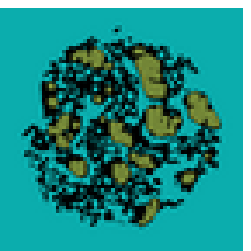}
\includegraphics[width=2cm]{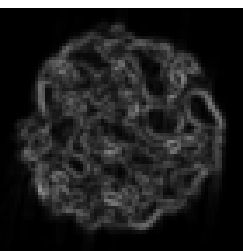}
\includegraphics[width=2cm]{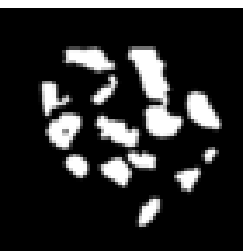}
\includegraphics[width=2cm]{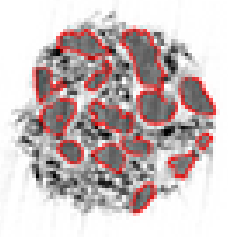}

\caption{Visualization of each step for the granular material A, B and C: the first image is the initial image, the second is the label localization inside the component, the third image (and the fourth image for the granular C because the complementary is composed by two components)  is the label localization inside each component of the complementary, the fourth image is the visualization of the double labels, the fifth image is the gradient image,  the sixth image is the catchment basin associated to the label included in the component, the seventh image is the foreground of the inner boundary of the previous image on the initial image.}
\label{fig:gew}
\end{center} 
\end{figure}

\begin{figure}
\begin{center}
\includegraphics[width=4cm]{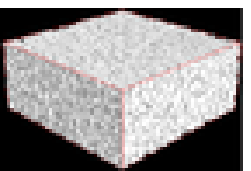}
\includegraphics[width=4cm]{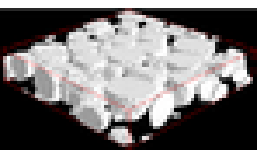}
\includegraphics[width=4cm]{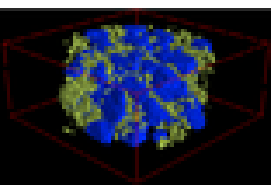}
\includegraphics[width=4cm]{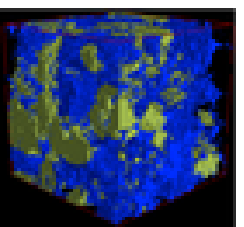}
\caption{The first image shows the white grains for the granular A. The second image shows the black grains for the granular B. The third image shows in yellow the small black grains and in blue the big grey grains for the granular C. The fourth image shows in blue the porosity (the black phase) and in yellow the black grains for the granular D. }
\label{fig:gew3d}
\end{center} 
\end{figure}

Two labels have to be localized: one included in the component and the other included  in the  component complementary.  Since the component complementary can be composed by more than one component, this procedure has to be operated on each complementary component (see figure~\ref{waterlabel}).
The procedure uses the tint information like the threshold segmentation: a threshold followed by a succession of morphological filters. The major enhancement of this approach is:  \textbf{it is not necessary that the label localization match the component} contrary to the threshold segmentation. It is just sufficient to have some labels in each connected component. The succession of morphological filter is \textbf{just an opening}. The opening operator has one parameter $k$ such as
 \begin{eqnarray*}
&A \bullet_k B = (A \oplus B_k) \ominus B_k\\
\mbox{with }& B_k = \overbrace{(B\oplus B)\ldots \oplus B}^{\mbox{times k}} 
\end{eqnarray*} 
 In this article, the selection of the threshold/opening parameters and are done manually following these constraints (see table~\ref{tab:param}):
\begin{enumerate}
\item the material specialist  checks if the visual segmentation matches the numerical segmentation,
\item if there is some experimental data about the volume fraction, we impose the correspondence between the experimental value and the numerical value obtained by segmentation.   
\end{enumerate}
This manual limitation is attenuated by a good property: some small parameters modifications  have no consequence on the final segmentation (see subsubsection~\ref{subsub:robustness}). So it is  easy to find the right parameters for a good segmentation because the range of the right parameters is large. This simple method gives some good results for the four granular materials. The figure~\ref{fig:gew} shows the different steps for the extraction of one component for the granular A, B and C. The figure~\ref{fig:gew3d} shows the 3D visualization of the multi-component extraction. In the next subsubsection, a method to evaluate the robustness is presented, in more this method  opens up the opportunity of an automatic evaluation of the parameters. 
\hspace*{-1cm}\begin{table}[h]
\centering
\begin{tabular}{|l|l|l|l|l|}
\hline
granular & component &label  &threshold &   opening \\
&  & & range & size \\
\hline
\multirow{2}{*}{A} & white grains &C & 150-255 & 1\\
  &&  B & 0-100 & 0\\
\hline
\multirow{2}{*}{B} &black grains &C & 0-80 & 1\\
 & & B & 160-255 & 1\\
\hline
\multirow{5}{*}{C}&black grains &C & 0-50  & 0\\
 & & B & 100-255& 0\\
&grey grains &C & 125-145 & 3\\
 & & B1 & 0-120& 2\\
 & & B2 & 160-255 & 1\\
\hline
\multirow{3}{*}{D} & porosity (black phase) &C & 0-100 & 0\\
 & & B & 150-255& 0\\
& black grains &C & 90-130 & 3\\
 & & B1 & 0-60& 1\\
 & & B2 & 170 & 0\\
\hline
\end{tabular}
\caption{B1 and B2 means the component number 1 and 2 of the component complementary. }
\label{tab:param}
\end{table}

\subsubsection{Robustness}
\label{subsub:robustness}
\begin{figure}
\begin{center}
\includegraphics[width=5cm]{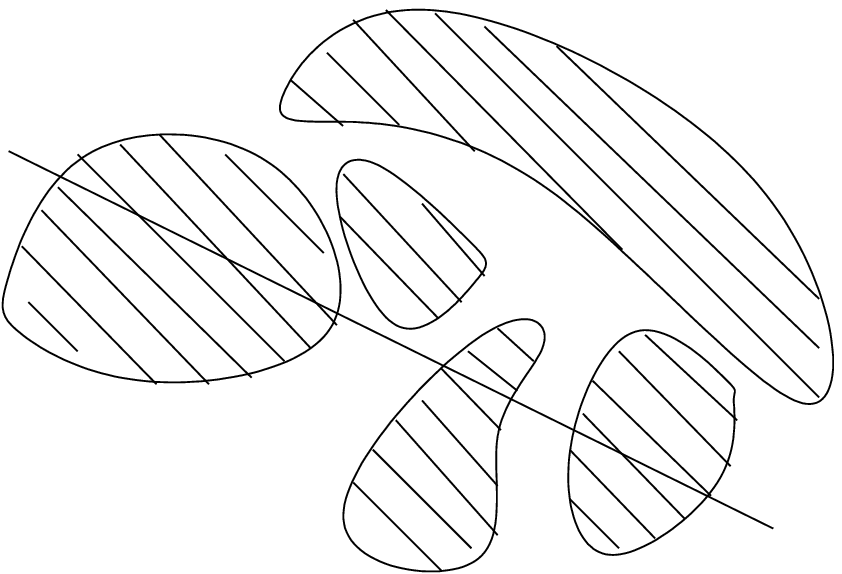}
\caption{A chord trough a granular material composed by two components}
\label{fig:chord}
\end{center} 
\end{figure}

\begin{figure}
\begin{center}
\includegraphics[width=5cm]{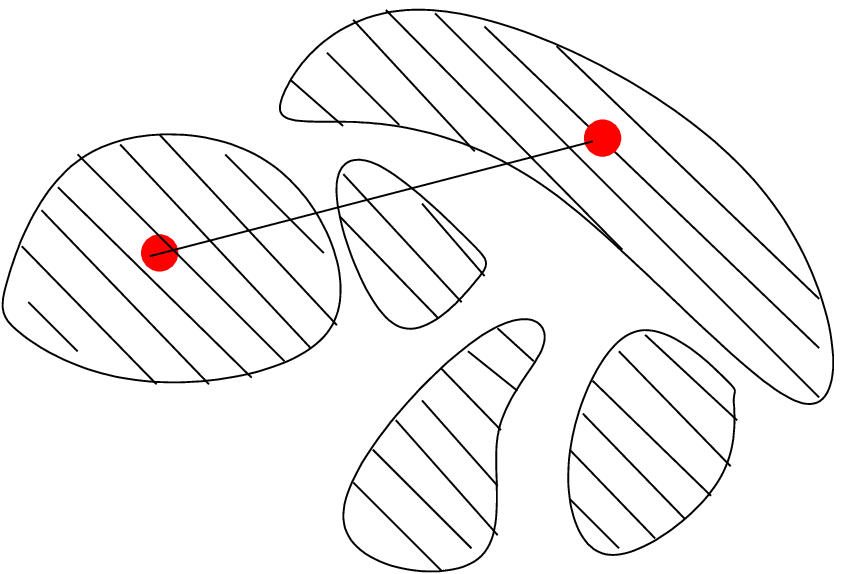}
\caption{A yardstick drawn randomly in a granular material composed by two components}
\label{fig:corr}
\end{center} 
\end{figure}

A definition of the segmentation robustness can be the stability of some morphological descriptors depending on the parameters fluctuation. This subsubsection has two paragraphs: morphological analysis and robustness evaluation.
\paragraph*{Morphological analysis}
The $\mu$-chord distribution function and the two-point correlation function provide a way to get a morphological analysis of the mesoscopic divided material and particularly for granular materials \cite{CICCARIELLO1981,Levitz1992,Porod}. These functions can be computed either in 2D or 3D assuming some general conditions similar to the Hadwiger conditions. Their determination gives information about volume fraction, average granular size, mean curvature, granular shape, surface roughness and structural correlation.\\
1) A chord is a segment belonging either to the grains or to the background and having its two extremities on the interface. As shown in Fig.~\ref{fig:chord}, $\mu$-chord are obtained by tracing random and homogeneous distributed straight lines through the microstructure. The chord-length distribution function gives the probability of getting a chord length between $r$ et $r+dr$, belonging either to the grain, $f_{g,\mu}(r)$,or to the background $f_{b,\mu}(r)$.\\
2) Let a yardstick drawn randomly in the material. The two-point correlation functions gives the probability of having  both extremities of the yardstick belonging to the grains, $f_{g,\mbox{2 point}}(r)$ or to the background $f_{i,\mbox{2 point}}(r)$ \footnote{For a biphasique material, the two point correlation functions of one phase is formulated in function of the two point correlation functions of the other phase.} (see figure~\ref{fig:corr}).\\

\begin{figure}
\begin{center}
\includegraphics[width=8cm]{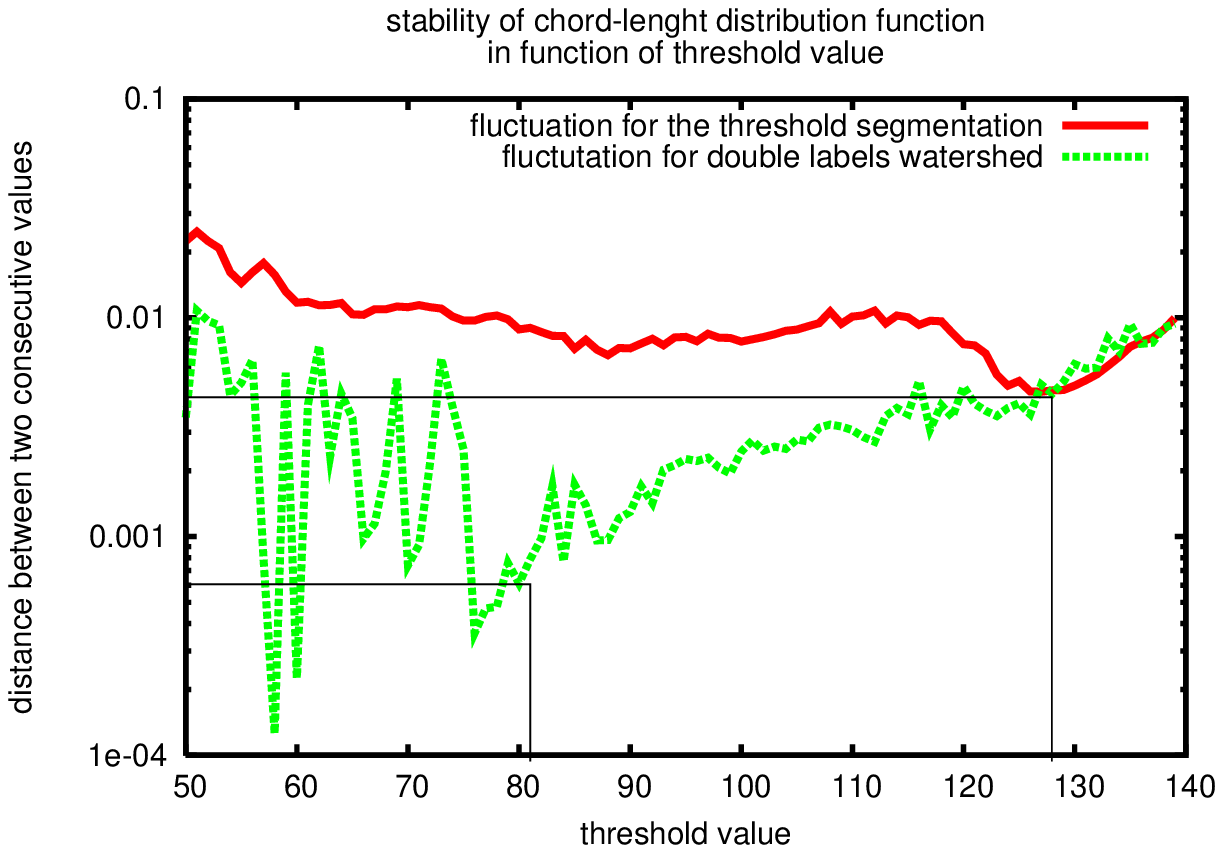}
\includegraphics[width=8cm]{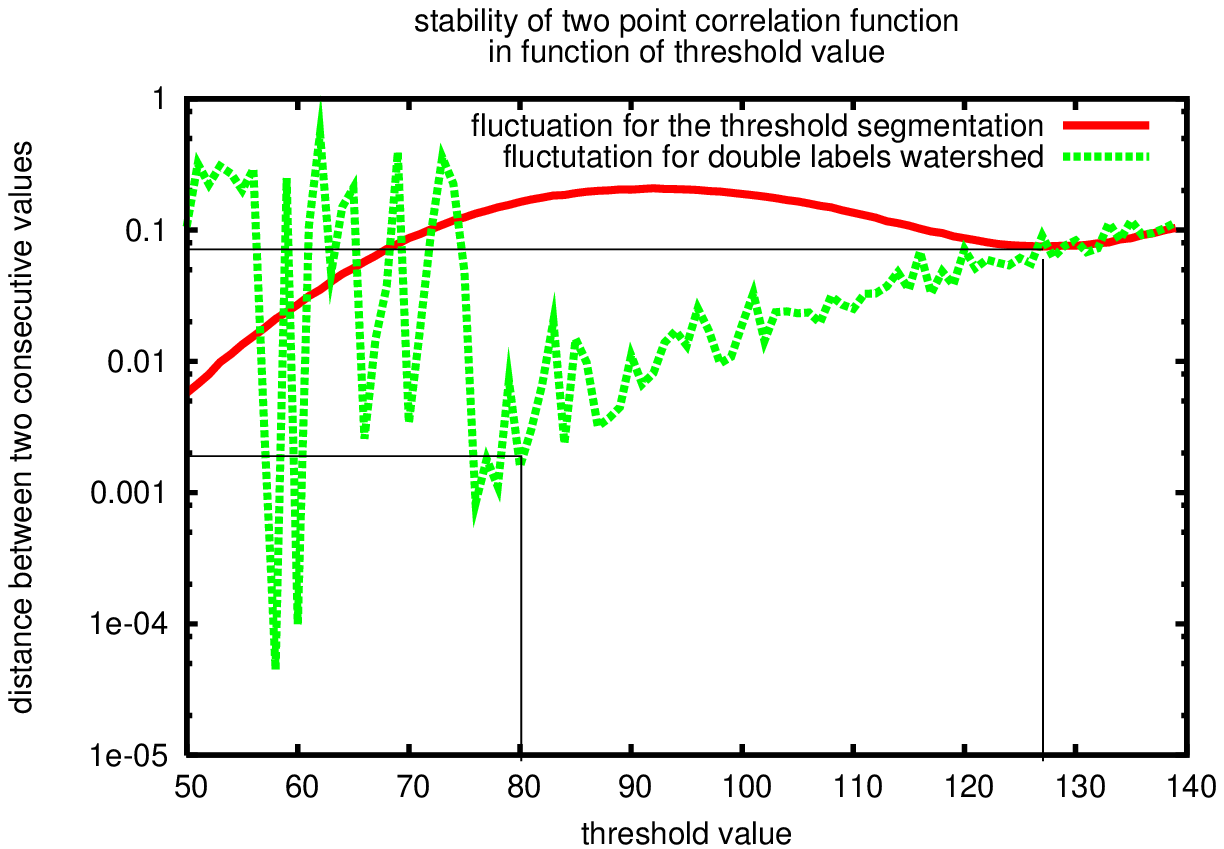}
\caption{Application for the granular material B: for the threshold segmentation, the threshold value is selected (the value 128 corresponds to the valley on the histogram)  and for the double labels watershed, the threshold value to localize the label inside the grains is selected (the value 90 has been chosen manually to give a result matching the visual segmentation). For  both distances, the double labels watershed is more stable of  one decade than the threshold segmentation.}
\label{fig:robust}
\end{center} 
\end{figure}

\begin{figure}
\begin{center}
\includegraphics[width=8cm]{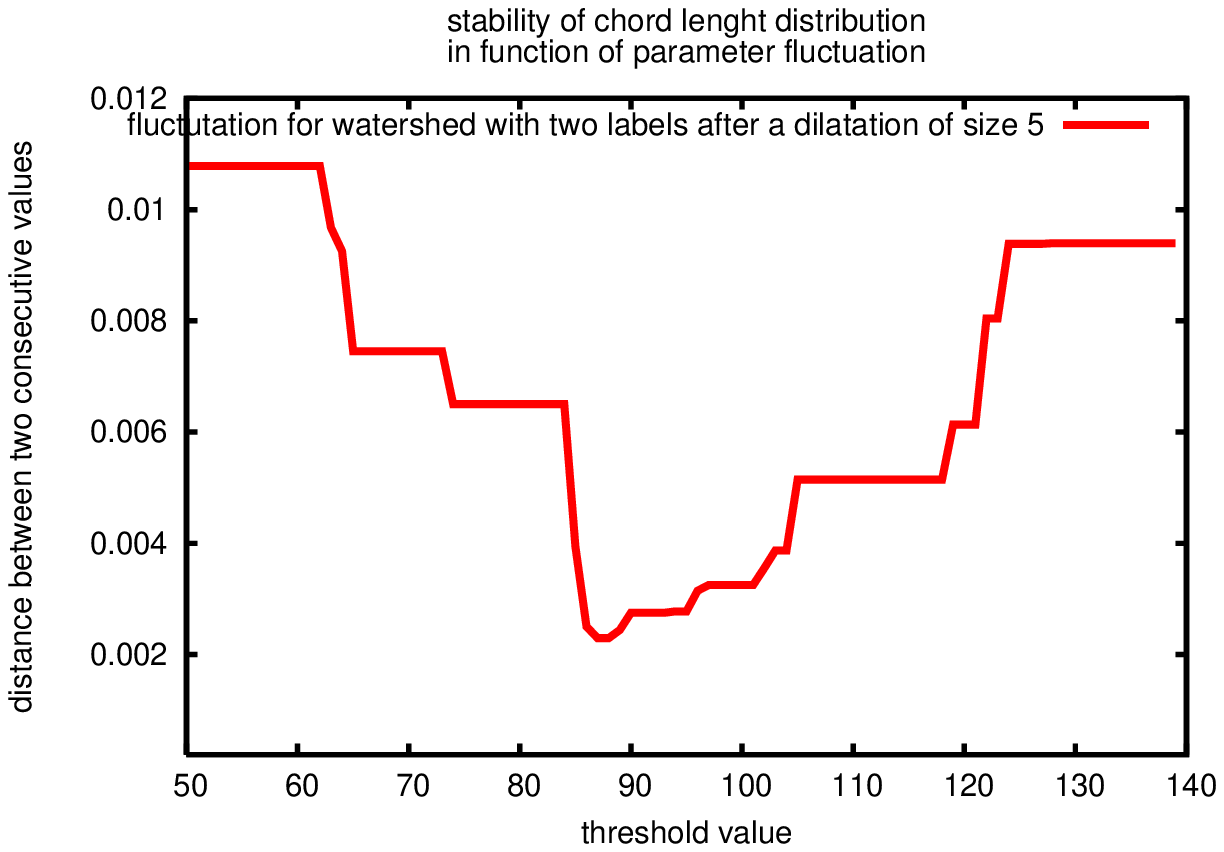} 
\includegraphics[width=8cm]{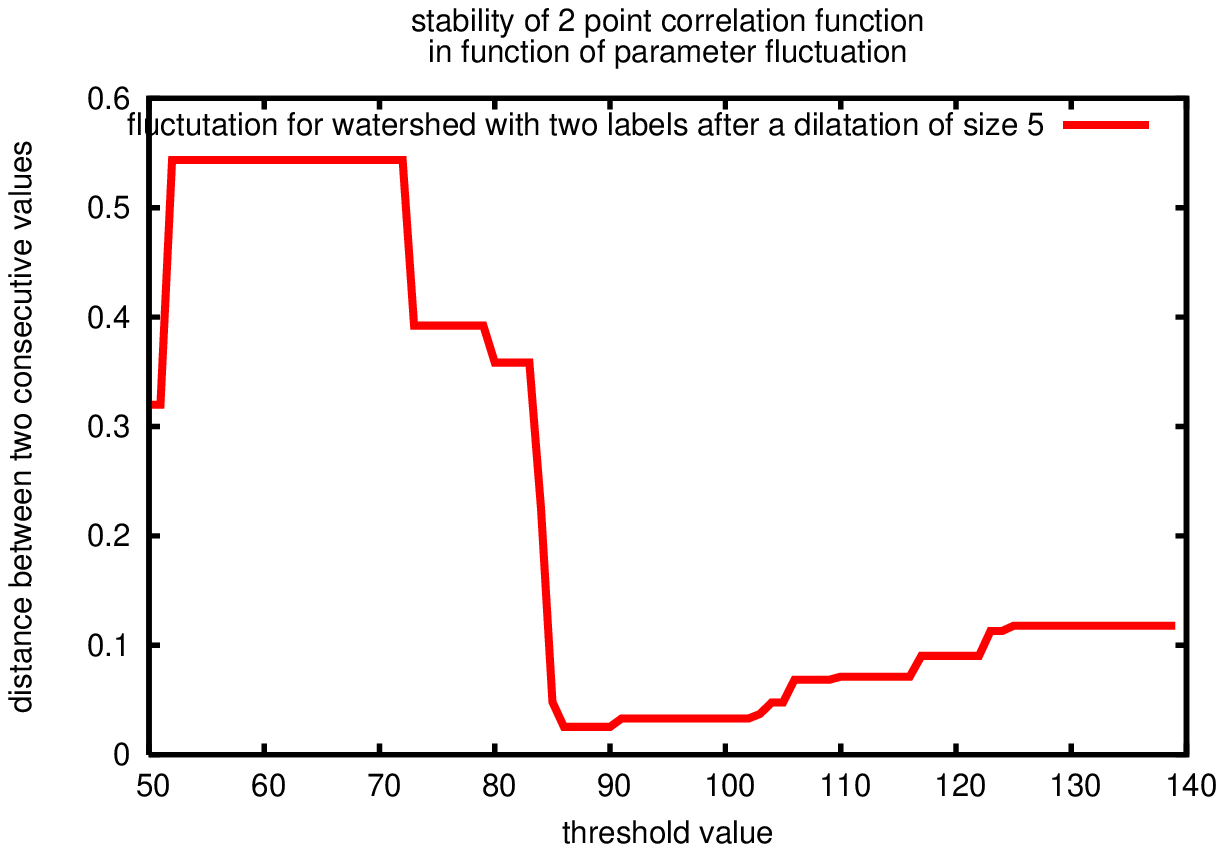}
\caption{After a dilatation of parameter 5 on the graph, the minimum on both curves is 87. This automatic value is closed that the manual value: 90.}
\label{fig:automatic}
\end{center} 
\end{figure}

\paragraph*{Robustness evaluation}
 Let $S_\Lambda(I)$ be the resulting binary image after the application of the threshold segmentation or the double labels watershed on the initial image $I$ with the parameters $\Lambda$. Let $\Lambda_b$ the parameters that gives the "best" binary image.\\
The method to check the stability of a segmentation method is:
\begin{enumerate}
\item  select a parameter, $\lambda$, in the parameters, $\Lambda$,
\item  starting from well below of $\lambda_b$, increase step by step the chosen parameter, $\lambda_{i} = \lambda_{i-1} + \epsilon $, until a value well above of $\lambda_b$, 
\item for each step,
\begin{itemize}
\item apply the segmentation procedure to get the binary image $S_{\Lambda_{i}}(I)$ associated to the parameter $\lambda_{i}$,
\item calculate the distance for the  two point correlation function  and for the chord-length distribution between $S_{\Lambda_{i}}(I)$ and $S_{\Lambda_{i-1}}(I)$ \footnote{The distance is: $\Vert f^{S_{\Lambda_{i}}(I)}_{g,\mbox{2 point or }\mu} - f^{S_{\Lambda_{i-1}}(I)}_{g,\mbox{2 point or }\mu }\Vert_2 $ }.
\end{itemize}
\end{enumerate}
The figure~\ref{fig:robust}) shows that the double labels watershed is more stable than the threshold segmentation. For the double labels watershed, plotting these curves will allow an automatic determination of the threshold value to localize the labels (see figure~\ref{fig:automatic}).
Whatever the parameters, there is always a problem: some pairs of grains, close one to each other, become connected. A split procedure will be applied to separate the connected grains. 

%% file: dynamic.tex
\section{Split the grains}
\begin{figure}
\begin{center}
\includegraphics[width=8cm]{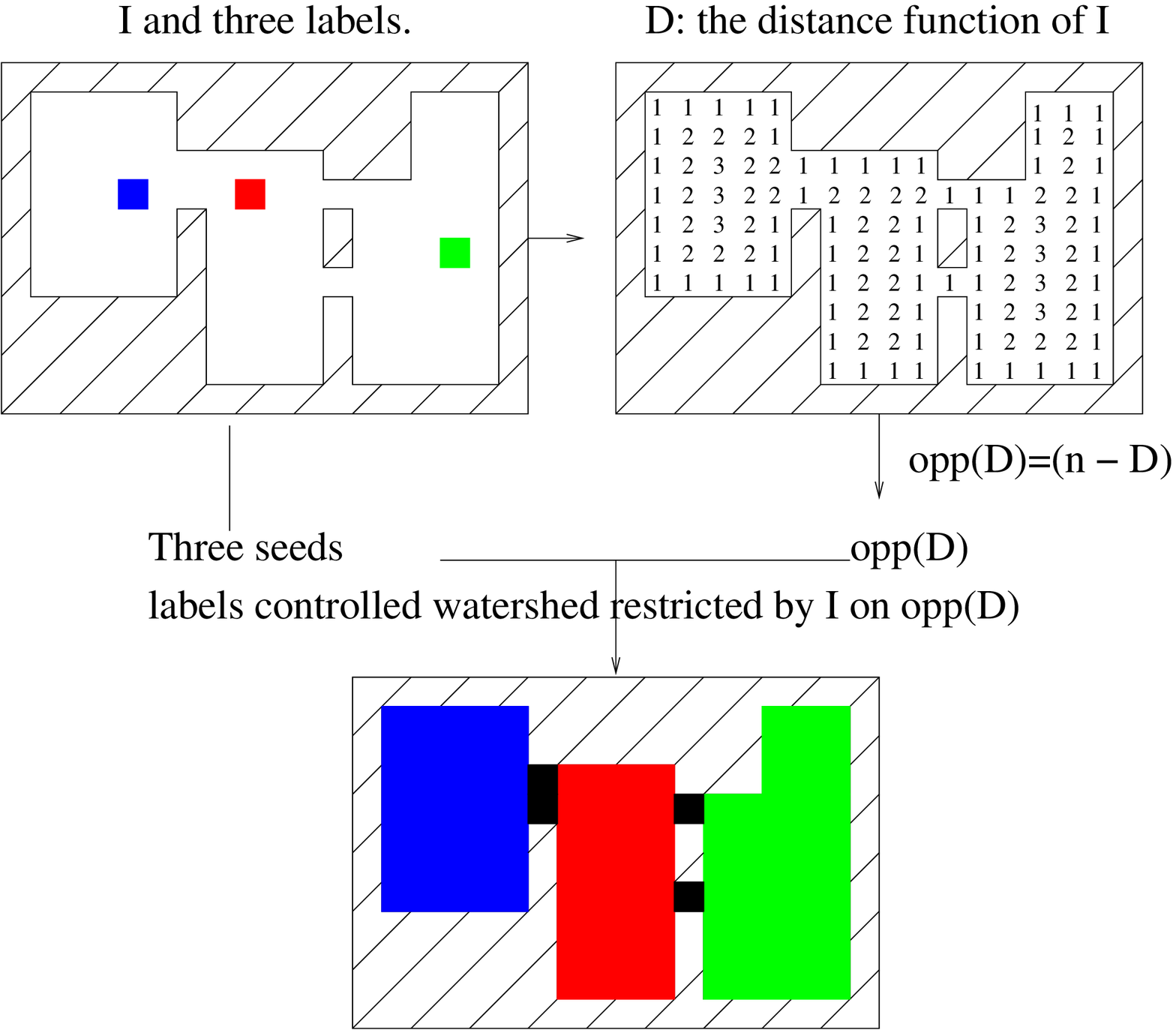}
\caption{Grains extraction: the watershed lines/surfaces are located on the narrow lines/surfaces because the narrow lines/surfaces are located on the crest lines of the opposite of the distance function of $I$.}
\label{poreprincipe}
\end{center} 
\end{figure}

\begin{figure}
\begin{center}
\includegraphics[width=10cm]{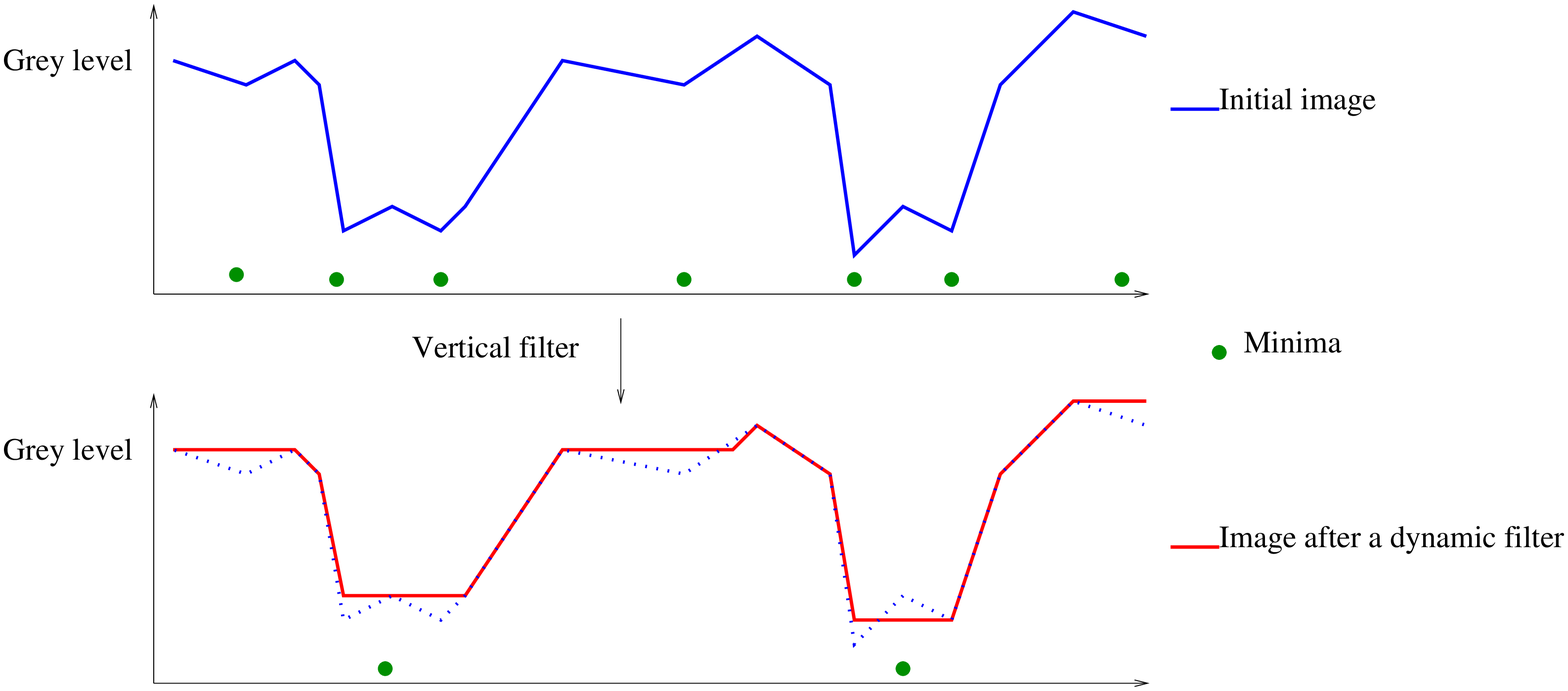}
\caption{The dynamic filter. Before the application of the dynamic filter, there are many regional minima (green bullets). After the application of the dynamic filter, there are only two regional minima.}
\label{dynamic}
\end{center}  
\end{figure}

\begin{figure}
\begin{center}
\includegraphics[width=4cm]{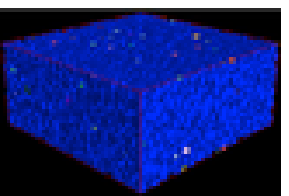}
\includegraphics[width=4cm]{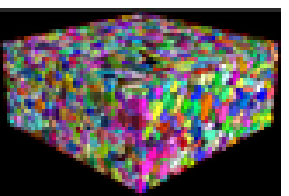}
\includegraphics[width=4cm]{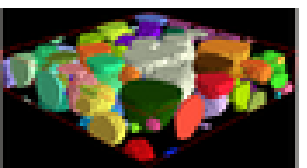}
\includegraphics[width=4cm]{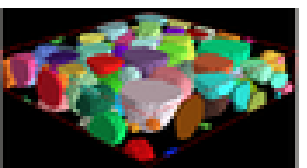}
\includegraphics[width=4cm]{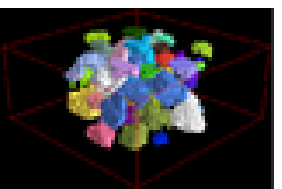}
\includegraphics[width=4cm]{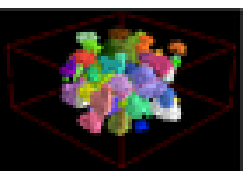}
\caption{Each colour represents a grain. For each serie, the first image is the visualization of the clusters on the segmented binary image and the second image is the catchment basins after the splitting procedure. The serie and the granular is 1-A, 2-B, 3-C. For each granular, the dynamic parameter is equal to 1.}
\label{dynav}
\end{center} 
\end{figure}

\begin{figure}
\begin{center}
\includegraphics[width=4cm]{image/bornert_visu_3d.eps}
\includegraphics[width=4cm]{image/bornert3dseg_split.eps}\\
\includegraphics[width=4cm]{image/iex_visu_3d.eps}
\includegraphics[width=4cm]{image/iex_seg_split.eps}\\
\includegraphics[width=4cm]{image/2211_visu_3d.eps}
\includegraphics[width=4cm]{image/22113dsplit.eps}\\
\includegraphics[width=4cm]{image/estaille_visu_3d.eps}
\includegraphics[width=4cm]{image/estaille3d_seg_both.eps}\\
\includegraphics[width=4cm]{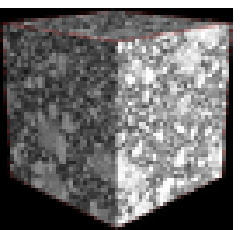}
\includegraphics[width=4cm]{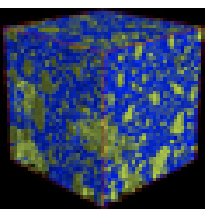}\\
\caption{Corresdance between the 3d grey-level image and the segmented image. The last serie is a cement paste material composed by three phases: porosity (in black), hydrate (in middle grey) and anhydrous grains (in light). In the segmented image, the porosity is in blue and the the anhydrous grains is in yellow.}
\label{fig:endseg}
\end{center} 
\end{figure}

 For the both segmentation procedure, the final result shows one difficulty:  some groups of two grains close one to each other become connected after the segmentation. So if a classical cluster procedure\cite{HOSHEN1976} is applied on the segmented binary image, these numerical connected grains are considered as a single grain. To overcome this problem, we apply a procedure introduced by L. Vincent in 1993\cite{Vincent1993}.\\
The algorithm principle is to apply the watershed transformation controlled by labels  restricted by the binary image, $I$, on the topographic surface that is the opposite of the distance function of $I$. In this approach, the crucial step is the labels localization. Each catchment basin is associated to a grain (see figure~\ref{poreprincipe}). \\
In a first approach, the labels can be the regional minima of the opposite distance function of the binary image. However, in practice, this merging produces an important over-extraction of grains due to local irregularities of the binary image shape, generating a set of uncontrolled and unwanted labels. To avoid this problem coming from too many labels, the opposite distance function f of the binary image, $O(dist(I))$, is filtered. A vertical filter is used in order to individualize each grain with a single label. The vertical filter uses the notion of dynamic with a parameter $h$. The operator is a geodesic erosion of the image $O(dist(I))+h$ under the image $O(dist(I))$ that fills the valleys with depth lower than $h$\cite{Grimaud1992,Tariel2008d} (see figure~\ref{dynamic}).
The splitting procedure has been successfully applied to the granular A-B-C\footnote{For the granular D, the physical property under investigation is the electrical resistivity and the segmented components image are sufficient to predict it.} (see figure~\ref{dynav}) because each grain of the granular phase has been correctly extracted.  
This method has been applied for granular materials but its extension to other materials is straightforward. The figure~\ref{fig:endseg} shows the correspondance between the 3d grey-level image and the segmented image for the material presented in this paper and also for a cement paste. 

%% file: SRGPA_segmentation_tariel_08.bbl
\begin{thebibliography}{10}

\bibitem{Bernard2005}
D.~Bernard.
\newblock 3d quantification of pore scale geometrical changes using synchrotron
  computed microtomography.
\newblock {\em Oil \& Gas Science And Technology-Revue De L Institut Francais
  Du Petrole}, 60(5):747--762, September 2005.

\bibitem{Beucher1991}
S.~Beucher.
\newblock The watershed transformation applied to image segmentation.
\newblock {\em Conference on Signal and Image Processing in Microscopy and
  Microanalysis}, pages 299--314, 1991.

\bibitem{Beucher1979}
S.~Beucher and C.~Lantuejoul.
\newblock Use of watersheds in contour detection.
\newblock In {\em real-time edge and motion detection}. International workshop
  on image processing, 1979.

\bibitem{CICCARIELLO1981}
S.~Ciccariello, G.~Cocco, A.~Benedetti, and S.~Enzo.
\newblock Correlation-functions of amorphous multiphase systems.
\newblock {\em Physical Review B}, 23(12):6474--6485, 1981.

\bibitem{DAVIS1975}
L.~S. Davis, A.~Rosenfeld, and J.~S. Weszka.
\newblock Region extraction by averaging and thresholding.
\newblock {\em Ieee Transactions On Systems Man And Cybernetics},
  SMC5(3):383--388, 1975.

\bibitem{Deriche1987}
R.~Deriche.
\newblock Using canny criteria to derive a recursively implemented optimal edge
  detector.
\newblock {\em International Journal Of Computer Vision}, 1(2):167--187, 1987.

\bibitem{Grimaud1992}
M.~Grimaud.
\newblock A new measure of constrast: dynamics.
\newblock In {\em Proc. SPIE Vol. 1769, pp. 292-305, Image Algebra and
  Morphological Processing III}, 1992.

\bibitem{Hall2008}
S.A. Hall, N.~Lenoir, G.~Viggiani, J.~Desrues, P.~Bésuelle, and Di~Michiel M.
\newblock Evolution of strain localisation in granular media: observation
  through in-situ xray micro-tomography and 3d-volumetric digital image
  correlation.
\newblock {\em in preparation}, 2008.

\bibitem{Han2008}
M.~Han, V.~Tariel, S.~Youssef, E.~Rosenberg, M.~Fleury1, and Levitz P.
\newblock The effect of the porous structure on resistivity index curves. an
  experimental and numerical study, 2008.

\bibitem{HOSHEN1976}
J.~Hoshen and R.~Kopelman.
\newblock Percolation and cluster distribution .1. cluster multiple labeling
  technique and critical concentration algorithm.
\newblock {\em Physical Review B}, 14(8):3438--3445, 1976.

\bibitem{Jeulin2007}
D.~Jeulin, T.~Clemenceau, M.~Faessel, V.~Tariel, G.~Contesse, and Fanget A.
\newblock Morphological analysis of 3d images of pyrotechnical granular
  materials obtained by microtomography.
\newblock In {\em 17 DYMAT Technical Meeting Cambridge}, 2007.

\bibitem{Lenoir2008}
N.~Lenoir, Y.~Pannier, S.A. Hall, Bornert M., P.~B\'esuelle, J.~Desrues, and
  G.~Viggiani.
\newblock Combining x-ray ct and 3d digital image correlation for studying
  strain localization in granular materials.
\newblock In {\em 8th International Workshop on Bifurcations and Degradations
  in Geomaterials, Lake Louise, Alberta, Canada.}, 2008.

\bibitem{Levitz1992}
P.~Levitz and D.~Tchoubar.
\newblock Disordered porous solids - from chord distributions to small-angle
  scattering.
\newblock {\em Journal De Physique I}, 2(6):771--790, June 1992.

\bibitem{PANDA1978}
D.~P. Panda and A.~Rosenfeld.
\newblock Image segmentation by pixel classification in (gray level, edge
  value) space.
\newblock {\em Ieee Transactions On Computers}, 27(9):875--879, 1978.

\bibitem{Porod}
G.~Porod.
\newblock Small angle x-ray scattering.
\newblock In {\em Syracuse}, 1965.

\bibitem{Rosenfeld1982}
A.~Rosenfeld and Kak A.C.
\newblock {\em Digital Picture Processing}.
\newblock Academic Press, New York, 1982.

\bibitem{Serra1982}
J.~Serra.
\newblock {\em Image Analysis and Mathematical Morphology - Vol. I . 610 p.}
\newblock Ac. Press, London, 1982.

\bibitem{Tariel2008c}
V.~Tariel.
\newblock Conceptualization of seeded region growing by pixels aggregation.
  part 2: how to localize a final partition invariant about the seeded region
  initialisation order.
\newblock {\em submitted}, 2008.

\bibitem{Tariel2008d}
V.~Tariel.
\newblock Conceptualization of seeded region growing by pixels aggregation.
  part 3: a wide range of algorithms.
\newblock {\em submitted}, 2008.

\bibitem{Tariel2008}
V.~Tariel, D.~Jeulin, A.~Fanget, and G.~Contesse.
\newblock 3d multi-scale segmentation of granular materials.
\newblock {\em Image Analysis \& Stereology}, 2008.

\bibitem{Vincent1993}
L.~Vincent.
\newblock Morphological grayscale reconstruction in image analysis:
  Applications and efficient algorithms.
\newblock {\em Ieee Transactions On Image Processing}, 2(2):176--201, April
  1993.

\bibitem{WESZKA1978}
J.~S. Weszka.
\newblock Survey of threshold selection techniques.
\newblock {\em Computer Graphics And Image Processing}, 7(2):259--265, 1978.

\end{thebibliography}
